# Forecasting the outcome of spintronic experiments with Neural Ordinary Differential Equations

Xing Chen[1,2], Flavio Abreu Araujo[3,4], Mathieu Riou[4], Jacob Torrejon[4], Dafiné Ravelosona[2], Wang Kang[1], Weisheng Zhao[1], Julie Grollier[4], and Damien Querlioz[2*]

[1]Fert Beijing Institute, MIIT Key Laboratory of Spintronics, School of Integrated Circuit Science and Engineering, Beihang University, Beijing 100191, China.
[2]Université Paris-Saclay, CNRS, Centre de Nanosciences et de Nanotechnologies, Palaiseau, France.
[3]Institute of Condensed Matter and Nanosciences, Université catholique de Louvain, Place Croix du Sud 1, Louvain-la-Neuve, 1348, Belgium.
[4]Unité Mixte de Physique, CNRS, Thales, Université Paris-Saclay, Palaiseau, France.
*email: damien.querlioz@c2n.upsaclay.fr

## ABSTRACT

Deep learning has an increasing impact to assist research, allowing, for example, the discovery of novel materials. Until now, however, these artificial intelligence techniques have fallen short of discovering the full differential equation of an experimental physical system. Here we show that a dynamical neural network, trained on a minimal amount of data, can predict the behavior of spintronic devices with high accuracy and an extremely efficient simulation time, compared to the micromagnetic simulations that are usually employed to model them. For this purpose, we re-frame the formalism of Neural Ordinary Differential Equations (ODEs) to the constraints of spintronics: few measured outputs, multiple inputs and internal parameters. We demonstrate with Spin-Neural ODEs an acceleration factor over 200 compared to micromagnetic simulations for a complex problem – the simulation of a reservoir computer made of magnetic skyrmions (20 minutes compared to three days). In a second realization, we show that we can predict the noisy response of experimental spintronic nano-oscillators to varying inputs after training Spin-Neural ODEs on five milliseconds of their measured response to different excitations. Spin-Neural ODE is a disruptive tool for developing spintronic applications in complement to micromagnetic simulations, which are time-consuming and cannot fit experiments when noise or imperfections are present. Spin-Neural ODE can also be generalized to other electronic devices involving dynamics.

## Introduction

By combining the spin and charge degrees of freedom, spintronics offers multiple functionalities that are exploited in industrial applications for sensing and memory storage[1–4], and are currently investigated for communications[5] and information processing[6–11]. The rich functionalities of spintronic devices stem from the intricate magnetic textures from which they are formed, and the complex dynamical modes that can be excited in these textures. Spintronic systems, which have typical dimensions between a few nanometers to micrometers, can indeed not be considered as formed of a single spin and feature a large number of hidden variables: all the local magnetizations in the device. These spin textures can be dynamically excited by a wealth of physical quantities: magnetic fields, electrical currents or voltages, temperature, and pressure, all giving rise to different responses.

Micromagnetic simulations, the mainstream approach for predicting the behaviors of spintronic devices, solve this underlying complexity by dividing the structures in nanometers-wide cells, and simulating the dynamics of each cell using the Landau–Lifshitz–Gilbert equation, including the local and non-local interactions between micromagnetic cells[12–16]. This technique, therefore, involves a considerable number of coupled differential equations and requires very long simulation times, easily reaching weeks in time-dependent experiments or in micrometer-scale devices. Beyond their long simulation time, micromagnetic simulations come with essential limitations. The simulations have to be re-executed from scratch when the input parameters of the template need to be modified. Also, micromagnetic simulations can almost never fit quantitatively the results of an experiment. In a real experiment, the geometry of a nanostructure is indeed always approximate, the material parameters can never be perfectly controlled and may possess specific structural inhomogeneities. Experimental results are also easily affected by the injection of noise, the details of the measurement setups, and unknown external factors, which are challenging to consider in the micromagnetic modeling process. A new tool that could accurately predict experiments, even when all these non-idealities are present, would be invaluable. For example, experiments in the field of neuromorphic spintronics[7,17,18] currently involve months-long experimental campaigns to optimize all the inputs of the systems, a development time that could be reduced radically with an appropriate modeling tool. In industry, the development of spin-torque magnetoresistive memory (ST-MRAM) also involves a considerable amount of micromagnetic simulations and experiments to optimize device parameters[19].

The progress of artificial neural networks provides an alternative road to simulate the behavior of spintronic systems like this one and predict the results of experiments. In recent years, machine learning has been used increasingly in physics, for example, for discovering new materials and for learning physical dynamics from time-series data.[20–24,24–30]. However, the power of artificial neural network has never been applied to model, fit and forecast the complex experimental behavior of solid-state nanocomponents. In this context, a recent type of neural network, Neural Ordinary Differential Equation (ODE), holds special potential for modeling physical nanodevices, as it is specialized to predict the trajectories of dynamical systems (Fig. 1c). Neural ODEs, initially introduced in[31], are ODE models $\dot{\mathbf{y}} = f_\theta(\mathbf{y},t)$, where the function $f$ is expressed by a neural network with parameters $\theta$, which instead of being defined explicitly, can be trained through supervised learning. Based on examples of the behavior of the system (training data set), through the algorithm of stochastic gradient descent, the machine learning process can identify $\theta$ values that allow the Neural ODE to reproduce the presented trajectory examples. Once the Neural ODE has been properly trained to reproduce the training data set, it becomes a proper ODE model of the system dynamics and can be used to make predictions of its behaviors in unseen new situations.

Unfortunately, in their current form, Neural ODEs cannot be applied to the simulation of spintronic systems and solid-state devices in general, due to two major challenges:

- Neural ODEs require measuring the evolution of all the system variables, whereas in experiments and most applications, a single physical quantity is typically measured.

- Neural ODEs are not designed for dealing with external time-varying inputs.

In this work, we solve these two issues and show that Neural ODEs can be trained, using a minimal set of micromagnetic simulations or experimental data, to predict the results accurately for other inputs and material parameters, including the noise and intrinsic details of the specific studied nanodevices. In the rest of the paper, we first explain how we modified Neural ODEs to adapt them to spintronics and solve the two issues. We benchmark in detail our results with micromagnetic simulations, showing that the Spin-Neural ODE can produce accurate results in a complex skyrmion-based reservoir computing simulation, with considerably reduced simulation times with regards to micromagnetic simulations (20 minutes compared to three days). Finally, we train Neural ODEs to predict the results of real experiments on nanometer-scaled spin-torque oscillators. These experiments would be impossible to model with micromagnetic simulations, as they would require hundreds of years of simulation. By contrast, Neural ODEs obtain particularly accurate results, including the noise inherent to the experiments.

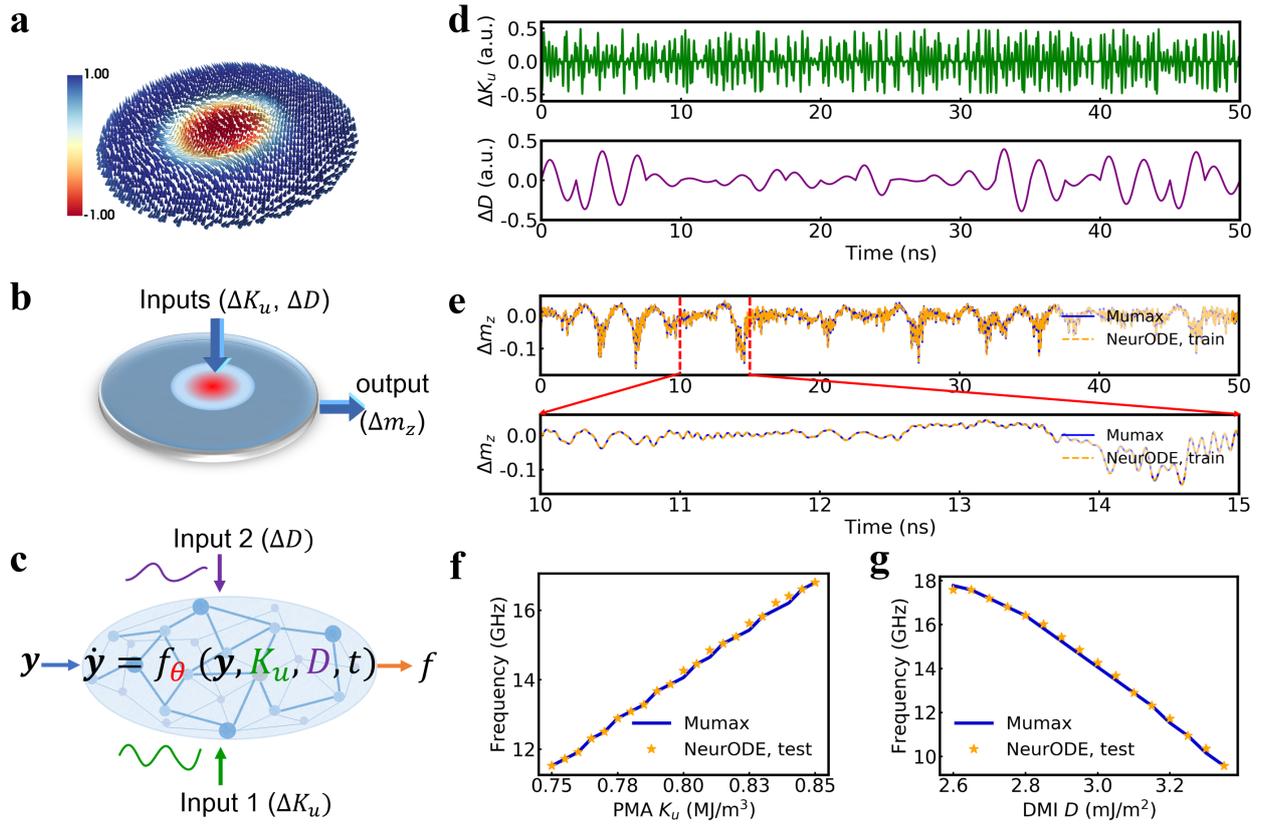

**Figure 1. Modeling a skyrmion-based device with micromagnetic simulation and Neural ODEs**. **a** Magnetic skyrmion configurations in a nano-disk. The color scale represents the out-of-plane component magnetization, and arrows denote the spin orientation. **b** Sketch of a device where a single skyrmion exists in the ferromagnetic layer. The behavior of the device depends on the Perpendicular Magnetic Anisotropy (PMA) constant (e.g., through VCMA effect) $K_u$ and the Dzyaloshinskii-Moriya Interaction (DMI) strength $D$. The output signal is the variation of perpendicular component of the mean magnetization $\Delta m_z$. **c** Sketch of a Neural ODE structure $\dot{\mathbf{y}} = f_\theta(\mathbf{y}, K_u, D, t)$ with $\Delta K_u$ and $\Delta D$ as external inputs into the neural network. $\mathbf{y}$ is a vector of system dynamics and $f$ is defined by a neural network (see Fig. 2 for details of the modeling method). **d** Time-dependent random sine variation $\Delta K_u$ ranging from -0.05 MJ/m$^3$ to 0.05 MJ/m$^3$ (corresponding to -0.5 to 0.5 in the graph) as an input (Input 1) and random sine variation $\Delta D$ ranging from -0.4 mJ/m$^2$ to 0.4 mJ/m$^2$ (corresponding to -0.4 to 0.4 in the graph) as another input (Input 2) applied to the skyrmion system. Equilibrium values of $K_u$ =0.8 MJ/m$^3$ and $D$ = 3 mJ/m$^2$ are used. **e** Predicted training output of $\Delta m_z$ by a Neural ODE in comparison with micromagnetic simulation (Mumax) results as a function of time. **f, g** Test results of the trained Neural ODE. The intrinsic response frequency of the skyrmion system for different values of $K_u$ in **f** and of $D$ in **g** calculated by using the trained Neural ODE (orange star) and by micromagnetic simulations (blue).

## Results

To introduce Spin-Neural ODEs, we consider a device made of a skyrmion, a chiral spin texture extensively studied today for its fascinating topological properties, as well as its stability, compact size, and non-volatility, all of high interest for applications[32–34] (Fig. 1a). We consider the device of Fig. 1b, with two inputs: the perpendicular magnetic anisotropy (PMA) constant $K_u$ and the Dzyaloshinskii-Moriya interaction (DMI) $D$. The output of the device is the the average magnetization perpendicular to the thin film axis, $\Delta m_z$, which translates directly to the electrical resistance of the device. In experiments, the PMA may be modulated by voltage through the VCMA effects, while the DMI is typically a constant of the material. However, to train a Neural ODE, we perform micromagnetic simulations where these two quantities vary artificially with random sine variations, during 50 ns, to explore the possible responses that the system can exhibit (Fig. 1d). Figs. 1e shows the elaborate variations that the output $\Delta m_z$ follows in these conditions. Our goal is to use this 50 ns time trace, which can be obtained using 40 minutes of micromagnetic simulations, to train a Neural ODE, able to predict the behavior of the system in any situation, including much longer simulations that would take days with micromagnetics.

### Extension of the Neural ODEs formalism to deal with incomplete information of dynamics

Neural ODEs take the conventional form of ordinary differential equations $\dot{\mathbf{y}} = f_\theta(\mathbf{y}, t)$, but where the function $f_\theta$ is a neural network (Fig. 1d). The vector $\theta$ contains the parameters of this neural network, i.e., its synaptic weights and neuron thresholds. The vector $\mathbf{y}(t)$ describes the different state variables of the system: the function $f_\theta$ is therefore a neural network that takes $\mathbf{y}$ as input and provides the derivative $\dot{\mathbf{y}}$ as output. Once an initial value of $\mathbf{y}$ is given, the system dynamics is computed automatically by calling an ODE solver. Training a Neural ODE model, i.e., optimizing the $\theta$ parameters, normally requires the knowledge of the evolution of all these state variables over a collection of demonstrative examples[31]. After the training process has been completed, the Neural ODEs can be used to predict unseen data.

This conventional technique for training Neural ODEs has strong limitations for predicting the behavior of physical systems. It is often impossible to know all the state variables relevant to the dynamics of a physical system. For example, in the spintronic structure of Fig. 1b, only the mean magnetization $\Delta m_z$ is known. It can be considered the "output" of our nanodevice and used as parameter $y_1$ within the Neural ODE. However, $y_1 = \Delta m_z$ results from complex magnetic configurations and dynamics that cannot all be determined experimentally. Additional parameters are necessary to describe this underlying dynamics, which may be represented by unknown internal variables: $\tilde{y}_2$ to $\tilde{y}_m$.

Here we develop a new scheme to train Neural ODEs in this context of real experiments where, in practice, the knowledge of the system is always limited. Our idea originates in the insight that it is possible to convert a set of the first-order differential equations in multiple variables into a single higher-order differential equation in one variable. For example, let us consider the case with a single hidden variable $\tilde{y}_2$,

$$\left[ \begin{array}{c} \dot{y}_1 \\ \dot{\tilde{y}}_2 \end{array} \right] = \left[ \begin{array}{c} a y_1 + b \tilde{y}_2 \\ c y_1 + d \tilde{y}_2 \end{array} \right]. \tag{1}$$

By substituting $\tilde{y}_2$ and its derivative $\dot{\tilde{y}}_2$ calculated from the first ODE into the second one, a second-order ODE with variable $y_1$ can be derived as $\ddot{y}_1 = (a+d)\dot{y}_1 + (bc - ad)y_1$, where $\tilde{y}_2$ no longer appears. This equation is equivalent to the following ODE in terms of $y_1$ and $y_2 = \dot{y}_1$

$$\left[ \begin{array}{c} \dot{y}_1 \\ \dot{y}_2 \end{array} \right] = \left[ \begin{array}{c} y_2 \\ (a+d)y_2 + (bc - ad)y_1 \end{array} \right]. \tag{2}$$

This simple derivation suggests that an appropriate way for training a Neural ODE of $m$ internal variables where only one variable $y_1$ is accessible is to train a Neural ODE where the state vector $\mathbf{y}$ is composed of $y_1$ and its $(m-1)^{\text{th}}$-order derivatives (see Supplementary Note 6 for a discussion in arbitrary dimension). The drawback of using higher-order derivatives is the sensitivity to noise of derivatives, resulting in a relative noise level much larger than in the original signal (see Supplementary Note 6). To make the best use of the original information and to avoid any preprocessing procedures, in this work, we employ several successive time-delayed states as an alternative to derivatives: we consider the input vector

$$\mathbf{y}(t) = (y_1(t), y_1(t + \Delta t_d), y_1(t + 2\Delta t_d), \ldots, y_1(t + (k-1)\Delta t_d)), \tag{3}$$

where $k$ is the dimension of the new vector, and $\Delta t_d$ denotes a single delay time. These time-delayed variables contain all the information provided by the high-order derivatives, but are less prone to noise. This scheme, which we introduced here in a qualitative manner, can also be justified mathematically by using a formalism known as the embedding theorem (see Supplementary Note 7).

## Extension of the Neural ODEs formalism to deal with time-varying external inputs

The second challenge for employing Neural ODEs for predicting the behavior of physical systems is to include time-varying external inputs, such as the anisotropy and the DMI change in Figs. 1b-d. In this case, the time derivative of the **y** variable is not only dependent on its current state, but also related to its current step's input, a situation that cannot be described in the traditional forms of Neural ODE.

Supplementary Note 6 shows how such inputs can be included in our approach. This note shows that, mathematically, an $m$-dimensional input-output ODE system, $\dot{\mathbf{y}} = f_\theta(\mathbf{y}, e(t), t)$ with $e(t)$ as input can be converted into an $m^{\text{th}}$-order ODE in the first variable $y_1$, depending on $e(t)$, but also the first to $(m-1)^{\text{th}}$-order derivatives of $e(t)$. Accordingly, with our method, the delay vector of equation 3 augmented with the extra input $(e(t), e(t+\Delta t_d), \ldots, e(t+(k-1)\Delta t_d))$ is used as input the $f_\theta$ function.

A system with multiple inputs can then be modeled by incorporating time-delayed versions of all inputs. As illustrated in Figs. 2a-b, we treat the time $t$ as an extended element of vector $\mathbf{y}(t)$ into the neural network and concatenate its time derivative, which is a constant one value, as a known output of the neural network. In this way, the external inputs at any moment can be chosen deterministically and given to the neural network. For a clearer visualization, the whole procedure of our technique is provided in Algorithm 1.

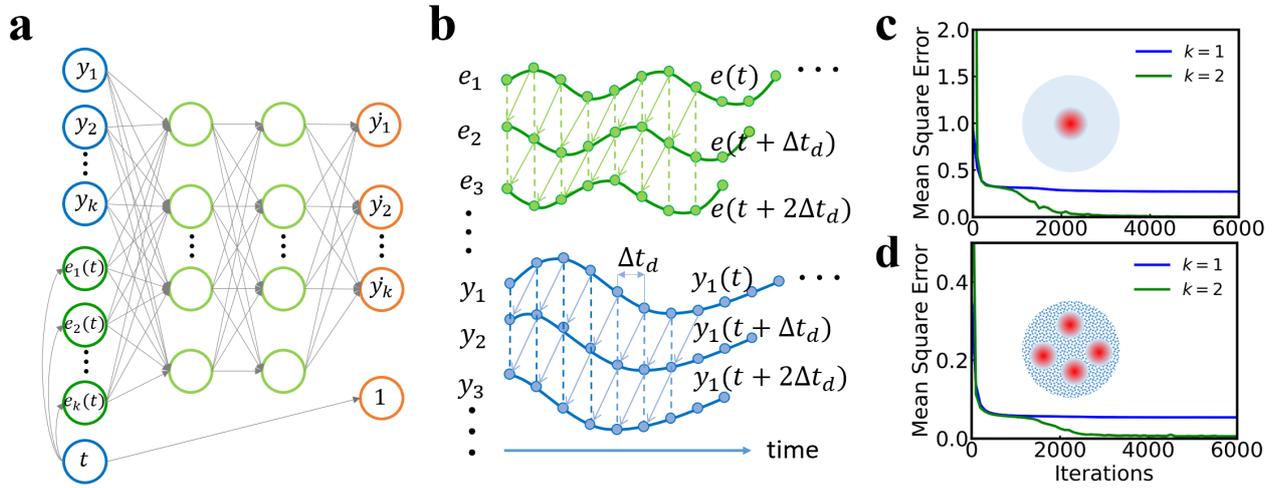

**Figure 2. Extending the Neural ODE formalism to predict spintronic results.** A wide range of dynamical systems can be modeled using ODEs, such as the simple pendulum motion, skyrmion-based devices, and spintronic oscillator dynamics. However, in real-world applications, the underlying physical dynamics are not always fully measurable, or accessible, which means that the dynamics of some hidden parameters in ODEs models are unknown. Our goal is to model a Neural ODE, $\dot{\mathbf{y}} = f_\theta(\mathbf{y}, e(t), t)$, where $f$ is defined by a neural network and $e(t)$ is the time-dependent input into the system, using the incomplete information of the system dynamics. **a** Schematic graph of the neural network ($f_\theta$) in a Neural ODE. The input to the neural network consists of two parts, one is the $k$ dimensional vector related to the observed system dynamics, where $y_1$ is the observed dynamics and $y_2$ to $y_k$ are the time delayed dynamics of $y_1$ (as shown in **b**), another part comprises the time-dependent external inputs, in which $e_1$ is the original input and $e_2$ to $e_k$ are the time delayed versions of $e_1$. The output of the neural network is a vector of time derivatives of the corresponding input system dynamics $(y_1, y_2, \ldots, y_k, t)$. Here, the derivative of the time variable $t$ is 1, which is determined as a prior knowledge. **b** Illustration of the time-domain system dynamics and external time-dependent input dynamics used for modeling the Neural ODEs. The blue curves are the system dynamics, where $y_1$ is the original observed trajectory, $y_2 = y_1(t+\Delta t_d)$ is one time step shifted of $y_1$, $y_3 = y_1(t+2\Delta t_d)$ is two time steps shifted of $y_1$, etc. Here, $\Delta t_d$ denotes the single time delay interval. The green curves are the external inputs, where $e_1 = e(t)$ is the original input dynamics, $e_2 = e(t+\Delta t_d)$ is one time step shifted of $e_1$, $e_3 = e(t+2\Delta t_d)$ is two time steps shifted of $e_1$, etc. Through the augmentation, the reconstructed system dynamics $(y_1, y_2, y_3, \ldots, y_k)$ containing the information of the unknown state variables can be used to train the Neural ODEs and then the trained Neural ODEs can be applied to make predictions for other inputs. **c, d** Training error (Mean Square error, MSE) as a function of iterations for $k=1$ and $k=2$ for a one-skyrmion system (**c**) and a multi-skyrmions system with grain inhomogeneity (**d**) with electric voltage as input through the VCMA effect.

**Algorithm 1:** Training Neural ODEs using incomplete system dynamics and external input

**input** : Time intervals $T = \{t_0, t_1, \ldots, t_{n-1}\}$ with uniformly spaced step $\Delta t$, time-dependent input $E = \{e(t_0), e(t_1), \ldots, e(t_{n-1})\}$, observed scalar output trajectory $Y = \{y(t_0), y(t_1), \ldots, y(t_{n-1})\}$, mini-batch time length $bt$, mini-batch size $bs$, iterations $N$, dimension of the new vector $k$ (number of delays $k-1$), a single time delay interval $\Delta t_d = \Delta t$, and Neural ODE parameters $\theta$ with forward function:

    **function** forward ($y$):
        $t \leftarrow y[k]$         ▷ Extract the last dimension of vector $y$.
        $\dot{y} \leftarrow (f_\theta(y[0:k-1], e(t)), 1)$         ▷ The derivative of time $t$ is constant 1.
    **return** $\dot{y}$

**output:** Updated $\theta$

**for** $iter = 1, \ldots, N_i$ **do**
    1. Randomly select mini-batch with the initial time $t_b = \{t_{b_0}, t_{b_1}, \ldots, t_{b_{bs-1}}\}$ ($b_i \in [0, n-bt], i \in [0, bs-1]$, $i$ is an integer), mini-batch targets
    $y_{\text{true}} = \{(y(t_b), y(t_{b+1}), \ldots, y(t_{b+k-1}), t_b), \ldots, (y(t_{b+bt-1}), y(t_{b+bt}), \ldots, y(t_{b+bt+k-2}), t_{b+bt-1})\}$, initial points
    $y_0 = (y(t_b), y(t_{b+1}), \ldots, y(t_{b+k-1}), t_b)$, external input (at time step $t_{b+i}$, $i \in [0, bt-1]$, $i$ is an integer)
    $e(t_{b+i}) = (e(t_{b+i}), e(t_{b+i+1}), \ldots, e(t_{b+i+k-1}))$.
    2. Call the Neural ODE solver and compute the predicted output trajectory $y_{\text{pred}}$ using current $\theta$.
    3. Update $\theta$ by taking an ADAM step on the mini-batch loss, which is defined as Mean Square Error (MSE) of $y_{\text{pred}}$ compared to $y_{\text{true}}$.
**end**

## Application of Neural ODEs to predict the behavior of skyrmion-based systems

We now test the validity of our approach in the single-skyrmion system of Fig. 1. We train a Neural ODE with dimension $k = 2$ employing a three-layer neural network $f_\theta$, each hidden layer featuring 50 neurons, using Algorithm 1 (see Methods) and the 50-nanosecond trajectory of Fig. 1d. Fig. 1e displays the predicted training output of $\Delta m_z$ by the Neural ODE in comparison with the prediction from micromagnetic simulations, showing outstanding agreement. To test the performance of the trained Neural ODE from a physical perspective, we next use it to predict the intrinsic response breathing frequency of the skyrmion system for different values of $K_u$ or $D$ (see Methods). The results, shown in Figs. 1f-g, again show excellent agreement with the results obtained from micromagnetic simulations.

To highlight the importance of the choice of the dimension of the Neural ODEs, Figs. 2c shows the training process, where the training error (mean square error, MSE) converges rapidly to zero for a Neural ODE of dimension $k = 2$, in comparison with a Neural ODE with $k = 1$ (i.e., without augmentation of delayed state) (in this Figure, the anisotropy was used as sole input). The corresponding time-domain training outputs are shown in Supplementary Note 1. We observed that in the absence of noise, a good model can be trained for any dimension $k \geq 2$. This result can be interpreted by the fact that the physical modeling of the system can be described by two variables: skyrmion radius and coordinate phase. For the modeling of noisy time series, a dimension of two is insufficient to train a good model, and higher accuracy can be obtained by increasing the number of delays (Supplementary Note 2).

Further results regarding the training performance in terms of the number of neurons $N_h$ in the hidden layer, the sampling interval $\Delta t$ of the trajectory, the dimension $k$ of Neural ODEs, and different optimization algorithms can be found in Supplementary Note 1. Concerning the choice of $\Delta t_d$, this parameter should not be too small, as there would be almost no difference between different elements in a vector, and not too large, as the neighboring states may lose correlations. However, our results showed that the training results do not depend too significantly on $\Delta t_d$, and therefore, we used the sampling interval as $\Delta t_d$ value.

We also validated our algorithm in more complex situations. Neural ODEs were able to predict the behavior of multi-skyrmions systems with grain inhomogeneities exhibiting a distribution of perpendicular magnetic anisotropy (PMA) $K_u$, and with voltage as input, (see Fig. 2d and Supplementary Fig. 1c). Neural ODEs also worked when electric current is used as an input, causing the skyrmion to rotate within the device (see Supplementary Note 4).

To summarize, skyrmions systems are usually modeled by time-consuming micromagnetic simulations, and developing faster models is a challenging task. Isolated skyrmion dynamics is often modeled by analytical equations that neglect the skyrmion deformation in confined systems. On the other hand, it is particularly difficult to model multi-skyrmions system, especially taking into account imperfections at the material level. Here, we showed that Neural ODEs can be trained to model these different situations.

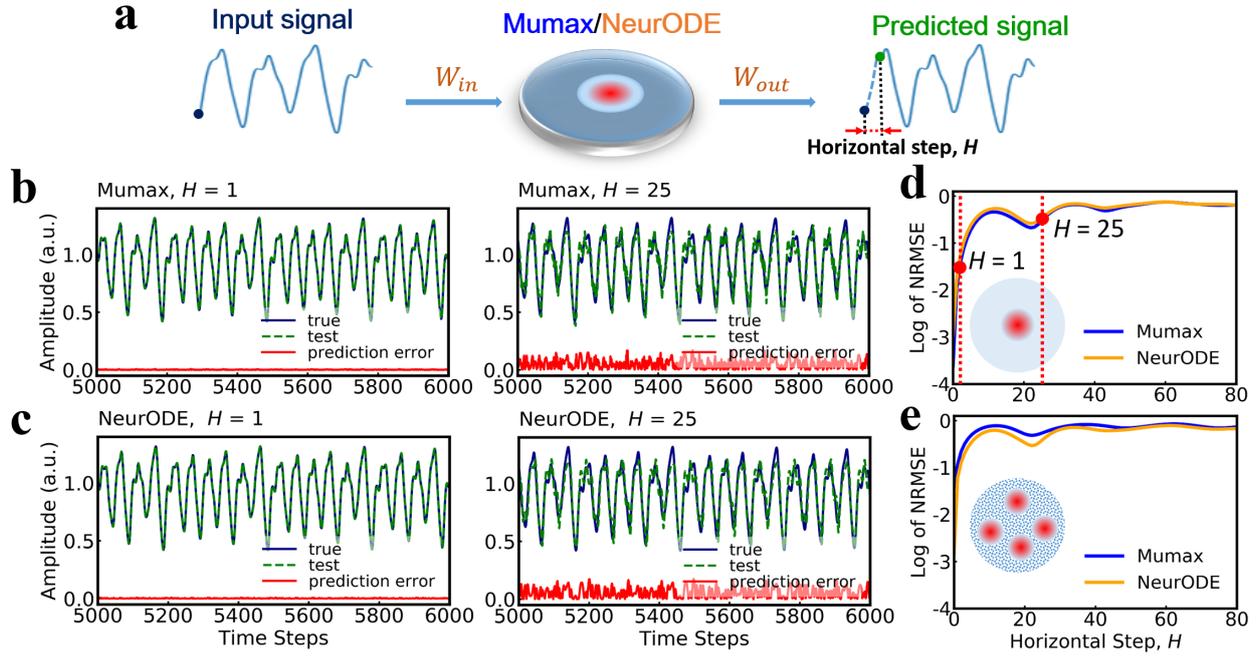

**Figure 3. Modeling of a sophisticated spintronic task, Mackey-Glass time series prediction with skyrmion systems, using Neural ODEs. a** Schematic graph of the procedure for doing the prediction task. The purpose is to predict the Mackey-Glass time series at a future time. The input signal, preprocessed through a read-in matrix $W_{in}$, is fed into the reservoir, which is a skyrmion system modeled by the trained Neural ODE or micromagnetic simulations. A trained output matrix $W_{out}$ is used for reading out the reservoir states and providing the predicted signal. **b-c** Selected testing (green dashed) results for prediction horizontal step $H = 1$ (for short term prediction) and $H = 25$ (for long term prediction), predicted by the one-skyrmion system modeled by micromagnetic simulations in **b** and the Neural ODE in **c**, in comparison with the true trajectory (blue) of Mackey-Glass time series. The red curves show the prediction error compared to the true trajectory. **d-e** Normalized root mean square error (NRMSE) as a function of prediction horizontal step $H$, in log scale, for the testing set by using the trained Neural ODE (orange) and micromagnetic simulations (blue) for the one-skyrmion system in **d** and the multi-skyrmions system in **e**.

## Benchmark test for Reservoir Computing: Mackey-Glass time series prediction

We now show that the Neural ODEs trained in the previous section can be used to predict the response of the spintronic system in a different setting, and with inputs that vary in a very different way, with computation time considerably reduced compared to micromagnetic simulations. We focus on a neuromorphic task called reservoir computing that exploits the intrinsic memory of complex dynamical systems, and apply it to the case of reservoirs made of single and multiple skyrmion textures[35–37]. The reservoir input corresponds to a chaotic time series (Mackey-Glass chaotic series, see Methods), and the goal of the task is to predict the next steps in the time series (Fig. 3a). The response of spintronic devices to such time series is particularly long to simulate with micromagnetic simulations. We simulated a reservoir computing experiment using the Neural ODEs trained in the previous section (which required 20 minutes of simulation time), as well as using micromagnetic simulations, as a control (requiring four days of simulation time). Figs. 3b and c show the time series predicted by a one-skyrmion system modeled by micromagnetic simulations and Neural ODEs, respectively, in comparison with the true trajectory (blue) of the Mackey-Glass time series. This data is presented in a situation where the skyrmion reservoir has to predict the next value in the Mackey-Glass time series ($H = 1$), and in a situation where it has to predict the value happening 25 steps later ($H = 25$), a much more difficult task due to the chaotic nature of the Mackey-Glass time series. In the $H = 1$ situation, the predictions of micromagnetic simulations and the Spin-Neural ODE match the true series perfectly, while for $H = 25$, a small prediction error happens, which appears consistent in both cases. To verify if the Spin-Neural ODE and micromagnetic simulations give equivalent predictions, Fig. 3d presents the accuracy of the prediction of the Mackey-Glass series, expressed in terms of Normalized Root Mean Square Error (NRMSE) as a function of horizontal prediction step $H$ (a lower value of the NRMSE means a more accurate prediction). We see that the NRMSE computed by the Neural ODE matches the one from micromagnetic simulations very precisely. Fig. 3e presents the same result for a multi-skyrmions system, where once again, the results of the

7/16

Neural ODE match those of micromagnetic simulations accurately. We also evaluated the prediction performance in terms of the number of virtual nodes of the reservoir, the time duration for each preprocessed input staying in the reservoir, and other techniques of training the output weights (see Supplementary Note 3).

    This demonstration of a demanding benchmark task exploiting the details of skyrmion dynamics indicates the potential of skyrmion system for reservoir computing, and highlights the quality of predictions by the Neural ODEs, with considerable improvement in computational efficiency (the Neural ODEs simulation were 200 times and 360 times faster than the micromagnetic simulations for the one-skyrmion system and multi-skyrmions system).

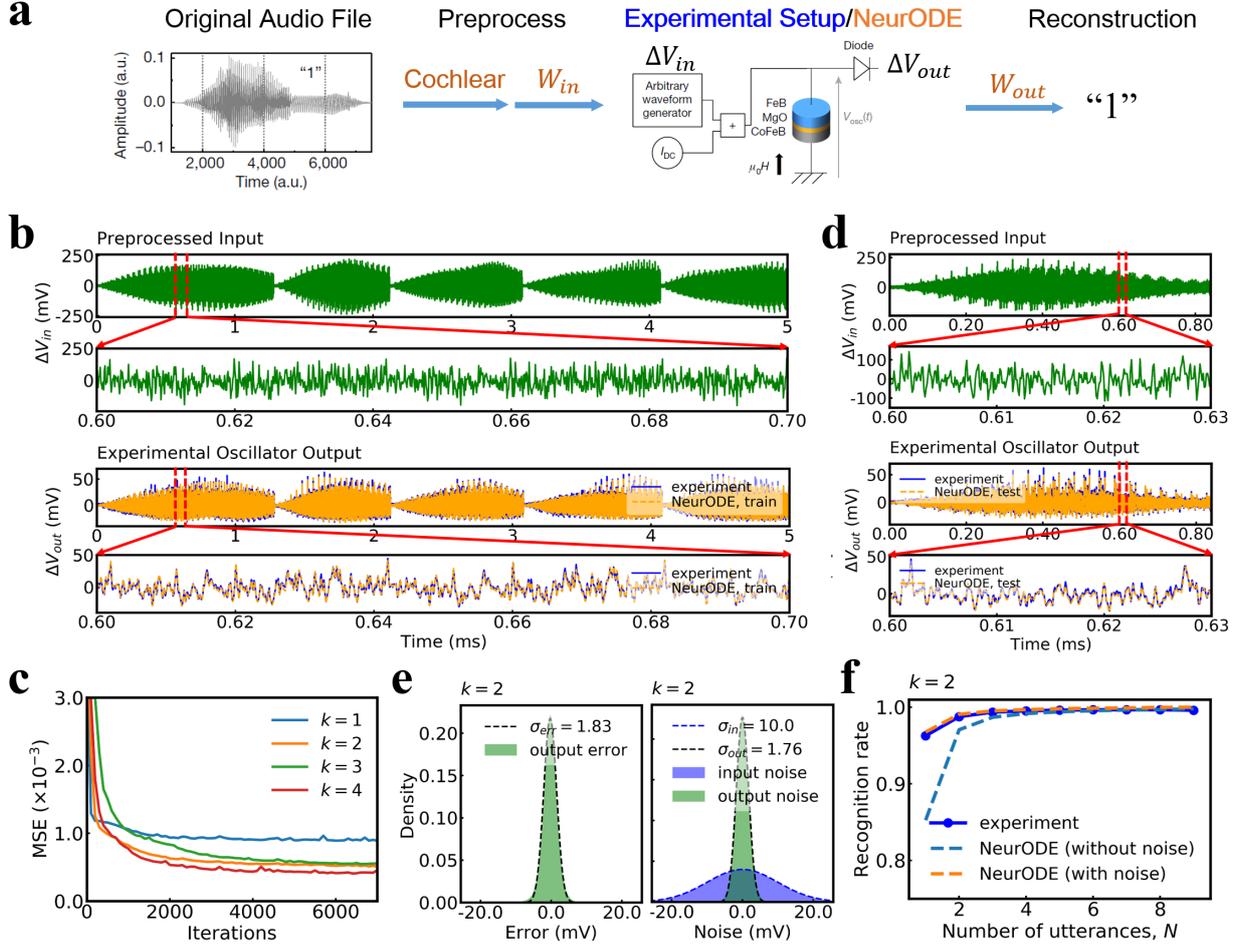

**Figure 4. Prediction of experimental results using Neural ODEs. a** Principle of the experiment. The original spoken digit in the audio waveform is preprocessed, by cochlear or spectrogram filtering, to form the preprocessed input into the oscillator. The output of the digit is reconstructed by reading out the recorded oscillator output through a trained matrix $W_{out}$ (see Methods). The purpose of modeling is to predict the experimental oscillator output given any preprocessed input $\Delta V_{in}$. This Figure shows the results obtained using cochlear filtering (results using spectrogram filtering are reported in Suppl. Fig. 7). **b** Training output trajectory of voltage $\Delta V_{out}$ predicted by Neural ODE (dashed orange) with corresponding preprocessed input $\Delta V_{in}$, in comparison with the experimental measurement (blue) for $k = 2$. A training set of 5-ms dynamics is adopted from the first utterance of the first speaker. A three-layer neural network $f_\theta$ with each hidden layer of 100 units is trained. **c** Training loss (MSE) of Neural ODE with $k = 1, 2, 3, 4$ as a function of iterations. **d** Selected response output signals predicted by the trained Neural ODE with corresponding preprocessed input, in comparison with the experimental output for digit eight of the third utterance of the third speaker. **e** Left to right: Error distribution (green shadow) and fitted Gaussian distribution (dashed curve) extracted by computing the difference between the output predicted by Neural ODE and the experimental measurement, a Gaussian noise (purple shadow) added in the preprocessed input into the Neural ODE and the corresponding noise distribution (green shadow) with fitted probability density function (pdf) (black dashed curve) in the predicted output trajectory solved by Neural ODE. $\sigma_{in}$ was adjusted so that $\sigma_{out} = 1.76$ mV is close to ($\approx$) $\sigma_{err} = 1.83$ mV. **f** Spoken digit recognition rate in the testing set as a function of utterances $N$ used for training. Because there are many ways to pick the $N$ utterances, the recognition rate is an average over all $10!/[(10-N)!N!]$ combinations of $N$ utterances out of the 10 in the dataset. The solid curve, blue dashed curve, orange dashed curve are the experimental result, Neural ODE result with noise considered, Neural ODE result without any noise, respectively.

**Predicting experimental measurements of nanoscale spintronic oscillator**

We now apply our approach to modeling real experimental data, obtained using the setup of[17]. This work showed experimentally that a nanoscale spintronic oscillator can be used as a reservoir computer to achieve a spoken digit recognition task (based on a principle similar to what we implemented in the skyrmion system). In this regime, the nano-oscillator is functioning as a nonlinear node to map the input signal into a higher dimensional space in which the input can be linearly separable (see Fig. 4a).

Modeling this experiment is a difficult challenge. Until now, analytical models of spintronic oscillators could reproduce experiments only qualitatively: it is challenging to construct a reliable model due to the high non-linearity of the devices as well as the impact of noise appearing in the experiments. Here, we firstly train a Neural ODE model of the oscillator dynamics by using only five milliseconds of experimental data (see Methods), then we use the trained model to predict the whole spoken digit recognition experiment of[17]. The results reported in Fig. 4 use cochlear preprocessing (see Methods).

Fig. 4b shows the five-millisecond trajectories used for training, as well as the result of trained Neural ODE of dimension $k = 2$, showing remarkable agreement. For a more quantitative assessment, Fig. 4c shows the training loss (MSE) of Neural ODEs with dimension $k$ ranging from one to four (a smaller MSE means a higher accuracy). The models with dimension two or greater reach a much smaller loss than a model of dimension one. This result is consistent with the physical intuition regarding these devices, which we would expect to be describable by coupled amplitude and phase equations, requiring therefore at least a two-dimension ODE[18]. It is also remarkable that the losses can be extremely close to zero, but not arbitrarily close. The impossibility of reaching zero loss can be attributed to the existence of noise in the experimental data, whereas the Neural ODE is entirely deterministic.

Next, the trained model can be utilized to predict the results of the spoken digit recognition experiment. Figs. 4b,d,e report results obtained using the Neural ODE of dimension $k = 2$. The aim of the task, described in Methods, is to classify the ten digits spoken by five different female speakers. Realizing this task experimentally involves a week-long experimental campaign, while it can be simulated in two hours using a trained Neural ODE. This task can also not be simulated by micromagnetic simulations, as it would require 716 years of simulation time on our reference GPU (this number was extrapolated based on the simulation of the dynamics of a nano-oscillator during one microsecond). Selected response output trajectories predicted by the trained Neural ODE and corresponding experimental output are shown in Figs. 4d-e, showing very good agreement.

We can now see how these results translate in terms of spoken digits recognition rate. Fig. 4g shows the recognition rate on spoken digits as a function of the number of utterances used. We see that the results obtained by a Neural ODE without noise do not match those of experiments. Incorporating noise within the Neural ODE is essential to predict the experimental data (this is particularly the case here, as reservoir computing is very susceptible to the disturbance of noise). To do this, we can rely on the five-millisecond training data. We extract the error distribution by computing the difference between the output trajectory predicted by Neural ODE and the experimental measurement and fit this error to a Gaussian law (Figs. 4f). We then inject Gaussian noise in the Neural ODE, as an additional input, with an amplitude chosen so that the output noise ($\sigma_{out}$) of the Neural ODE matches the standard deviation of the data ($\sigma_{err}$) of Fig. 4f (see Methods). When using the Neural ODE augmented with noise to simulate the spoken digits experiments, the digit recognition rates now match the experimental data very closely (Fig. 4g), making the Neural ODE augmented with noise a powerful tool to predict long and complex spintronic experiments.

Interestingly, the noise plays a role of suppressing the over-fitting of the output states from reservoir, actually improving the recognition rate. Conversely, we saw that task performance deteriorates with the injection of noise if the data has been preprocessed with the spectrogram filtering method, indicating that the output states from reservoir is under-fitted. This difference arises because the preprocessing procedure also contributes to the nonlinear transformation of input signal, and the cochlear method of preprocessing provides more nonlinearity than that of the spectrogram method[17] (more information about the impact of noise, and in particular in the spectrogram situation, is provided in Supplementary Note 5). It is remarkable that the Neural ODE augmented with noise is able to predict so subtle behaviors, which we were not able to realize from experimental data only.

## Discussion and Related Works

Our approach allows learning the underlying dynamics of a physical system from sample time-dependent data. Currently, important research relies on deep neural networks to predict results in physics. They are used to find abstract data representations[20], recover unknown physical parameters[22], or discover the specific terms of functions[23–25]. Other research uses recurrent neural network-based models[26–28] to learn and make predictions. These methods usually incorporate specific prior physical knowledge such as molecular dynamics[24,26], quantum mechanics[20], geospatial statistics[28], or kinematics[25] to help their models train faster or generalize. Moreover, few of these discrete models managed to include the relevant driving series to make predictions. Neural ODEs hold many advantages over the conventional neural networks used in these works: backpropagation occurs naturally by solving a second, augmented ODE backward in time; stability is improved with the use of adaptive numerical integration methods for ODEs; constant memory cost can be achieved by not storing any intermediate quantities of the forward pass; continuously-defined dynamics can naturally incorporate data which arrives at arbitrary times. However, until our work, there remained two challenges to apply Neural ODEs for predicting the behavior of physical systems, e.g., not only it is often impossible to acquire all the dynamics of state variables of the system, but also external input that will affect the system dynamics should be considered in many real-world applications.

Our work addresses both issues, and before ours, other works have attempted to solve the first issue. One way is to introduce the inductive bias via the choice of computation graphs in a neural network[25,38–44]. For example, by incorporating the prior knowledge of Hamiltonian mechanics[43,44] or Lagrangian Mechanics[40–42] into a deep learning framework, it is possible to train models that learn and respect exact conservation laws. These models were usually evaluated on systems where the conservation of energy is important. Similarly, another strategy to deal with a dataset with incomplete information is through augmentation of original dynamics[45–47]. For instance, by augmenting the second-order dynamics of a system, one can learn the low-dimensional physical dynamics of the original system. However, nearly all the proposals mentioned above require the knowledge of additional dynamical information, such as higher-order derivatives, or extra processing of the original low-dimensional dynamics, which is not appropriate for dealing with noisy time series. Additionally, until our work, Neural ODEs-based methods for modeling time series had only been tested in a few classical physical systems, such as the ideal Mass-Spring system, Pendulum and Harmonic oscillators. Our work is the first one to apply Neural ODEs to predict the behavior of nanodevices, by resolving the above issues.

Most impressively, our method achieves significant improvement in time efficiency compared to the conventional simulation platforms. For example, to run the Mackey-Glass prediction task, it takes only 20 minutes for the trained Neural ODE, whereas it takes around three days and five days for micromagnetic simulations to predict the one skyrmion system and the multi-skyrmions system, respectively. To model the spintronic oscillator dynamics using real experimental data, output time series of a duration of 5 ms is sufficient for training a full Neural ODE model, able to predict system dynamics with any input. Essentially, constructing a reliable and accurate model is not the only purpose of Neural ODEs, they can be a strong support for fast evaluation, verification, and optimization of experiments. In this way, our work also paves a way to Neural ODE-assisted or machine learning-integrated simulation platform development.

In conclusion, we have presented an efficient modeling approach for physical ODE-based systems. The training data can be a single observed variable, even if the system features higher-dimensional dynamics. We have shown that the method can not only be applied for modeling ideal simulation data, and is also remarkably accurate for modeling real experimental measurements including noise. The trained model shows outstanding improvement (hundreds of times faster) in computation efficiency compared to standard simulation platforms of micromagnetic simulations. We have shown that Spin-Neural ODE is a strong support for making experimental predictions and dealing with more complicated computation tasks, such as the task of Mackey-Glass time-series predictions and spoken digit recognition in reservoir computing. The proposed method is a promising tool to bridge the gap between modern machine learning and the traditional way of researching, and could be applied to a variety of physical systems.

## Acknowledgements

This work was supported by European Research Council Starting Grant NANOINFER (reference: 715872) and BIOSPIN-SPIRED (reference: 682955). X.C. also acknowledges the support from the China Scholarship Council (No. 201906020155). W.K also acknowledges the National Natural Science Foundation of China (61871008), Beijing Natural Science Foundation (Grants No. 4202043), and the Beijing Nova Program from Beijing Municipal Science and Technology Commission (Z201100006820042). F.A.A. is a Research Fellow of the F.R.S.-FNRS. The authors would like to thank B. Penkovsky, L. Herrera Diez, A. Laborieux, T. Hirtzlin, and P. Talatchian for discussion and invaluable feedback.

## Author contributions statement

D.Q. directed the project. X.C. performed all the simulations with the help and advice from D.Q., J.G., F.A.A., M.R., W.K., D.R and W.Z. F.A.A., M.R., J.T. and J.G. provided the experimental measurement data of spintronic oscillators. All authors participated in data analysis, discussed the results and co-edited the manuscript.

## Conflict of Interest Statement

The Authors declare no competing interests.

# Methods

## Micromagnetic simulations

Our micromagnetic simulations are performed in the MuMax3 platform (abbreviated to Mumax in the main text)[48], an open-source GPU-accelerated micromagnetic simulation program. The default mesh size of 1 nm × 1 nm × 1 nm is used in our simulations. The following material parameters are adopted: exchange stiffness $A = 15$ pJ/m, saturation magnetization $M_s = 580$ kA/m, damping constant $\alpha = 0.01$, interfacial DMI strength $D = 3.5$ mJ/m$^2$, and default PMA constant of the ferromagnetic layer $K_u = 0.8$ MJ/m$^3$. In addition, we set the VCMA coefficient $\xi$ as 100 fJ·V$^{-1}$m$^{-1}$ based on some recent experiments[49,50]. Here, the typical thickness of the insulating layer is 1 nm. Under these conditions, with an applied voltage of 0.1 V (an electric field of 0.1 V/nm), the PMA constant in the ferromagnetic layer will change by 10 kJ/m$^3$.

For the simulation of single-skyrmion dynamics with voltage input, a nanodisk with a diameter of 80 nm is used. The external input voltage to the system is random sine voltage with a frequency of 4 GHz and with amplitude ranging from -2 V to 2 V (corresponding to variation of PMA value $\Delta K_u$ from -0.2 to 0.2 MJ/m$^3$, see Supplementary Figure 1). For the multi-skyrmions system with grain inhomogeneity, the diameter of the nanodisk is 120 nm, and the grain size is 10 nm. Random 20% PMA variation, random 20% DMI strength variation, and 5% random cubic anisotropy direction variation are applied. The external input voltage to the system is a random sine voltage with a frequency of 4 GHz and with amplitude ranging from -2 V to 2 V (corresponding to variation of the PMA value $\Delta K_u$ from -0.2 to 0.2 MJ/m$^3$).

For the parameters-based simulations in Fig. 1, the diameter of the nanodisk is 100 nm. For the training set, the external input $\Delta K_u$ is a random sine with a frequency of 4 GHz and with amplitude ranging from -0.05 to 0.05 MJ/m$^3$, fluctuated around 0.8 MJ/m$^3$. The external input $\Delta D$ is a random sine with a frequency of 0.4 GHz and with amplitude ranging from -0.4 to 0.4 mJ/m$^2$, fluctuated around 3.0 mJ/m$^2$ (see Fig. 1d). The perpendicular average magnetization variation $\Delta m_z$ of the system is recorded every $p = 2.5$ ps as output. For the testing set, to get the response frequency of each material value of $K_u$ ($D$), we firstly supply a pulse with amplitude $\Delta K_u = 0.04$ MJ/m$^3$ ($\Delta D = 0.1$ mJ/m$^2$) lasting for 1 ns, then the magnetization variation $\Delta m_z$ is recorded. Finally, a Fourier transform is conducted on the output trajectory of $\Delta m_z$ to obtain the frequency. Simulation time of 37 mins, 41 mins, and 43 mins for 50 ns dynamics are needed for the training set of the one skyrmion system, multi-skyrmions system, and parameters-based system simulations.

## Training method of Neural ODE

To train the Neural ODE, we build a training set consisting of a collection of sample trajectories $y_{\text{true}}$, sampled at multiple time steps with a time interval $\Delta t$. We use the mean squared error (MSE) between these points and the corresponding trajectories predicted by the Neural ODE $y_{\text{pred}}$ over all time steps as the "loss function", i.e., the value that the training process aims at minimizing. To achieve the minimization of the loss, the gradients of the loss with respect to the parameters $\theta$ are computed through a technique called adjoint sensitivity method[31], then the $\theta$ parameters can be updated by using gradient descent optimization algorithms (usually stochastic gradient descent or Adaptive Moment Estimation[51]), until the MSE approaches zero. In this work, we train a Neural ODE in the form of $\dot{y} = f_\theta(y, e(t), t)$ in which $y(t) = (y_1(t), y_1(t + \Delta t_d), y_1(t + 2\Delta t_d), \ldots, y_1(t + (k-1)\Delta t_d))$ and $e(t) = (e(t), e(t + \Delta t_d), e(t + 2\Delta t_d), \ldots, e(t + (k-1)\Delta t_d))$ with Adaptive Moment Estimation.

The number of training data points $n$ = 10,000, 15,000, 10,000, and 50,000 are used for the one skyrmion system, the multi-skyrmions system, the parameter-based system and experimental data of oscillator, respectively. A sampling interval $\Delta t$ for training is determined according to the original recorded output period $p$. Specifically, $p = 2.5$ ps, 100 ns are used for Mumax simulations and experimental measurements, respectively. In principle, a Neural ODE system can be properly modeled as long as the training data are continuous in time. Considering the trade-off between the accuracy of the model and the time efficiency for processing, $\Delta t = 2p$ is chosen for modeling the simulation data from modeling of data and $\Delta t = p$ for modeling the experimental data of oscillator (see training results with $\Delta t = p, 2p, 4p, 5p$ in Supplementary Note 1). The $f_\theta$ function of the Neural ODEs is a three-layer feedforward neural network. Each hidden layer features 50 units for Mumax data modeling, and 100 units for modeling of experimental data. The activation function is the tanh function except for the output layer. The values of the weights are initialized from the standard normal distribution with a mean of 0 and a standard deviation of 0.1. To train the Neural ODE using the Algorithm 1, we set the mini-batch time length $bt = 20$ and mini-batch size $bs = 50$. We use the optimization algorithm Adaptive Moment Estimation (Adam) with a learning rate 0.001 to update the hidden weight with loss gradients of MSE. Fixed step fourth-order Runge-Kutta method with 3/8 rule is used as the Neural ODE solver.

During the training, time intervals $T$ are normalized by a multiplying a factor of $\delta = 0.0125/p$ as default. In addition, normalized input and output is used for training. For the skyrmion-based model, the normalized input is shown in Fig. 1d: the variation of magnetization output $\Delta m_z$ is multiplied by ten for training. For modeling the experimental data of oscillator, both $\Delta V_{in}$ and $\Delta V_{out}$ are multiplied by ten for training.

**General concept of Reservoir Computing**

Reservoir Computing (RC) is a computational framework derived from recurrent neural network models and suited for temporal/sequential data processing[52]. A reservoir is a network of interconnected nonlinear nodes with feedback. It transforms non-linearly its input signal into a higher-dimensional space, so that the resulting signal can be linearly separable through a simple readout to a desired output. The key benefit behind RC is that only the output layer from the reservoir states is trained using a simple linear regression mechanism, such as ridge regression, as all connections inside the reservoir are kept fixed. RC thus brings great advantage for hardware implementation, because the computational power of naturally available systems, such as a variety of physical systems, substrates, and devices, can be leveraged for such computation tasks.

In general, there are several requirements for efficient computing by a physical reservoir. First, high dimensionality ensures the mapping from inputs signal into a high-dimensional space through a nonlinear transformation, so that the originally inseparable inputs can be separated in classification tasks, and the spatiotemporal dependencies of inputs can be extracted in prediction tasks. Second, the fading memory (or short-term memory) property is necessary so that the reservoir state is dependent on the current inputs and recent past inputs. Such a property is particularly important for processing temporal sequential data in which the history of the states is essential.

In the Mackey-Glass prediction task, a skyrmion system can emulate a reservoir when it is in transient dynamics under an external input of a varying current or voltage. Similarly, in the spoken digit recognition task, a single nonlinear oscillator is used as a reservoir by leveraging its current-induced transient response. In both cases, an additional preprocessing step, where the original input is multiplied by a mask, usually a random matrix, is needed to enable the virtual nodes to be interconnected in time. The output states from the reservoir are reconstructed in a similar way, where the states from the virtual nodes are read consecutively in time.

**Mackey-Glass prediction task**

In a chaotic system, small perturbations can result in radically different outcomes. The prediction of a chaotic system is thus a problematic task. As a chaotic system, the Mackey-Glass equation is generated from a delay differential equation (DDE),

$$\frac{dx(t)}{dt} = \frac{\beta x(t-\tau)}{1+x^{10}(t-\tau)} - \gamma x(t), \qquad (4)$$

where $x(t)$ is a dynamical variable, $\beta$ and $\gamma$ are constants. Chaotic time series can be achieved with $\beta = 0.2, \gamma = 0.1$ and $\tau = 17$[53].

In the main paper, our goal is to predict the Mackey-Glass time series at a future time step: the preprocessed input signal at the current time is fed into the reservoir, which maps it nonlinearly into higher-dimensional computational spaces (see Fig. 3a), and a trained output matrix $W_{out}$ is used for reading out the reservoir states. The number of steps between the future time step and the current step is defined as prediction horizontal step $H$. $W_{out}$ is different for the different $H$ values.

More precisely, the dataset for the prediction task is prepared in the following way. First, Eq.4 is solved for 100,000 integration time steps with $dt = 0.1$. Before the data are processed by the reservoir, they are downsampled with a downsampling rate of 10 to remove the possible redundancy in the input data[53]. Thus, we obtain 10,000 data points in total. The first $5,000 + H$ data points are used for the training, and the rest $5,000 - H$ are for testing. The first stage of the masking procedure is a matrix multiplication $W_{in} \cdot M_o$, where $W_{in} \in \mathbf{R}^{N_r \times 1}$ is the mask matrix with data values drawn from a standard normal distribution and $M_o \in \mathbf{R}^{1 \times L}$ is the original input data. Here $L = 5,000 + H$ is the number of the scalar input data points and $N_r$ is the reservoir size. We adopt $N_r = 50$ in the main text. As a consequence of the masking, we obtain the data matrix $M_e = W_{in} \cdot M_o \in \mathbf{R}^{N_r \times L}$. Then $M_e$ is column-wise flattened into a vector $e \in \mathbf{R}^{L \cdot N_r}$ and then fed into the reservoir of skyrmion systems.

Each of the value from $e$, multiplied by a voltage of 1.6 V ($\Delta K_u = 0.16\text{MJ}/\text{m}^3$) for the one skyrmion system and a voltage of 1 V ($\Delta K_u = 0.1\text{MJ}/\text{m}^3$) for the multi-skyrmions system, is provided as preprocessed input into the reservoir (modeled by a trained Neural ODE or Mumax) for $t_{step} = 4p = 10$ ps to make sure there is an effect of the input on the reservoir dynamics. In the following, the reservoir dynamics is recorded for every $t_{step}$ to form a vector of $M_y \in \mathbf{R}^{L \cdot N_r}$, which is then unflattened into a response matrix $M_x \in \mathbf{R}^{N_r \times L}$ for output reconstruction. We use the matrix $A \in \mathbf{R}^{N_r \times T_r}$ consisting the first $T_r = L - H$ columns of $M_x$ for training the read out Matrix. The teaching matrix $B \in \mathbf{R}^{1 \times T_r}$ consisting the last $T_r$ of the original signal $M_o$ is the time series to be predicted. A read out matrix $W_{out}$ is therefore constructed through the method of ridge regression,

$$W_{out} = (A \cdot A^T + \mu I)^{-1}(A \cdot B^T), \qquad (5)$$

where $\mu = 10^{-4}$ is used as regularization parameter. To evaluate the performance of the trained matrix $W_{out}$, NRMSE is calculated on the prediction results of the testing set $y_{pre}$ compared to the true trajectory of MG series $y_{tar}$,

$$\text{NRMSE} = \sqrt{\frac{1}{n_s \sigma_{tar}^2} \sum_{i=0}^{n_s} (y_{tar}(i) - y_{pre}(i))^2}. \qquad (6)$$

## Spoken digit recognition task

In the task of spoken digits recognition, the inputs are taken from the NIST TI-46 data corpus. The input consists of isolated spoken digits said by five different female speakers. Each speaker pronounces each digit ten times. The original input signals of the spoken digits are preprocessed using two different filtering methods: spectrogram and cochlear models. In both methods, firstly, each word is broken into $N_\tau$ time intervals of duration $\tau$. Here, $N_\tau$ can be different for different speakers. Then in each interval $\tau$, a frequency transformation is performed to convert the signal into the frequency domain with $N_f$ channels. This frequency signal is then processed by multiplying a mask matrix $W_{in}$ containing $N_f \times N_\theta$ random binary values for each interval to obtain $N_\theta \times N_f$ values in total as input to the oscillator. The number of virtual neurons is $N_\theta = 400$. Each preprocessed input value is consecutively applied to the oscillator as a constant current for a time interval $\theta = 100$ ns. For the classification task, the response matrix $S$ consisting of the output of all neuron responses for all of the $N_\tau$ intervals from $N$ utterances of ten digits of five speakers is used for training. The target matrix $Y$ contains the targets for each interval, which is a vector of 10 with the appropriate digit equaling to 1 and the rest equaling to 0. The output matrix $W_{out}$ is constructed by using the linear Moore–Penrose method,

$$W_{out} = YS^\dagger, \tag{7}$$

where † denotes the pseudo-inverse operator. For the evaluation on the testing set of the remaining (10-$N$) utterances, the ten reconstructed outputs corresponding to one digit are averaged over all of the time intervals of $\tau$ of one word, and the digit is identified by taking the maximum value of the ten averaged reconstructed outputs. The recognition rate is obtained by calculating the word success rate. For the recognition rate of each $N$, there is $10!/(N!(10-N)!)$ different ways to pick the $N$ training set; therefore, we average the results from all the different ways to obtain the final recognition rate (cross-validation). The experimental details, the preprocessing and post-processing procedures for the spoken digit recognition task can be found in[17].

## Experimental measurements on spintronic oscillator

The experimental implementation for the spoken digit recognition task is illustrated in Fig. 4a. The preprocessed input signal is generated by a high-frequency arbitrary-waveform generator and injected as a current through the magnetic nano-oscillator. The sampling rate of the source is set to 200 MHz (20 points per interval of time $\theta$). The bias conditions of the oscillator are set by a direct current source ($I_{DC}$) and an electromagnet applying a field ($\mu_0 H$) perpendicular to the plane of the magnetic layers. For the cochlear method, $I_{DC} = 7$mA, $\mu_0 H = 448$ mT. For the spectrogram method, $I_{DC} = 6$mA, $\mu_0 H = 430$ mT. See[17] for more details.

## Prediction of experimental data

We use the output signal recorded for every $p = \theta = 100$ ns from the oscillator. The first 50,000 output data points, which corresponds to a time length of 5 milliseconds, from the oscillator of the first utterance of the first speaker and corresponding preprocessed signal as input is used as training set to train a three-layer neural network $f_\theta$ with each hidden layer of 100 units. The trained model is then utilized to predict the response output of the oscillator of all other speakers. The trained $f_\theta$ function is a deterministic function without noise. We therefore evaluate the effect of the noise on the task performance by adding the noise drawn from a Gaussian distribution into the preprocessed input, so that the standard deviation of noise in the output trajectory predicted by Neural ODE ($\sigma_{out}$) is close to the standard deviation of error between experiments and the results of the noiseless trained ODE ($\sigma_{err}$), over the 5 millisecond training dataset, as shown in Fig. 4(f) for the cochlear method. See Supplementary Note 5 for the spectrogram method.

## Simulation machine specifications

For micromagnetic simulations, we used an Nvidia GeForce GTX 1080 graphics processor unit. For Neural ODE simulations, we used an Intel Xeon E5-2640 CPU with 2.5 GHz base clock frequency and 3.0 GHz maximum turbo frequency.

## Data and Code Availability

The micromagnetic simulations are performed using the freely available MuMax3 platform. Neural ODEs simulations are performed using Pytorch. The source codes used in this work are freely available online in the Github repository: https://github.com/Xing-CHEN18/NeuralODEs_for_physics

# Supplementary Information: Forecasting the outcome of spintronic experiments with Neural Ordinary Differential Equations

Xing Chen[1,2], Flavio Abreu Araujo[3,4], Mathieu Riou[4], Jacob Torrejon[4], Dafiné Ravelosona[2], Wang Kang[1], Weisheng Zhao[1], Julie Grollier[4], and Damien Querlioz[2*]

[1]Fert Beijing Institute, MIIT Key Laboratory of Spintronics, School of Integrated Circuit Science and Engineering, Beihang University, Beijing 100191, China.
[2]Université Paris-Saclay, CNRS, Centre de Nanosciences et de Nanotechnologies, Palaiseau, France.
[3]Institute of Condensed Matter and Nanosciences, Université catholique de Louvain, Place Croix du Sud 1, Louvain-la-Neuve, 1348, Belgium
[4]Unité Mixte de Physique, CNRS, Thales, Université Paris-Saclay, Palaiseau, France.
*email: damien.querlioz@c2n.upsaclay.fr

## Supplementary Note 1: Training performance for the skyrmion systems, using voltage as input

This note provides additional results on the training process of Neural ODEs on the one-skyrmion system without grain inhomogeneity and the multi-skyrmions system with grain inhomogeneity, using voltage as input. Fig. S1a shows the normalized random sine voltage input used in the training process. Figs. S1b-c show the predicted training output of $\Delta m_z$, by Neural ODE (orange dashed curve) and micromagnetic simulation output (blue curve), for the one-skyrmion and the multi-skyrmions system, respectively. Excellent agreement is seen.

Fig. S2 shows the training loss (MSE) of the one-skyrmion system without grain inhomogeneity as a function of iterations, for:

- **a** different numbers of units $N_h$ in hidden layers. We observe that, in general, the greater $N_h$ is, the faster the training loss converges.

- **b** different sampling intervals $\Delta t$ of the initially observed trajectory $y_1$ (the unit $p$ is equal to 2.5 ps, as in the main body text). These results show that an accurate Neural ODE model can be trained even if the original continuous time series is downsampled, until a certain point in which the neighboring states of the downsampled trajectory lose correlations.

- **c** different dimensions $k$ of the Neural ODE (with $k-1$ being the number of delays). The Neural ODE model can be successfully trained as long as $k \geq 2$. Generally, the higher the dimension is, the longer time it takes for the model to train.

- **d** different choices of training optimization algorithms. It is shown that the ADAM and stochastic gradient descent (SGD) methods ensure a quick convergence of the loss function.

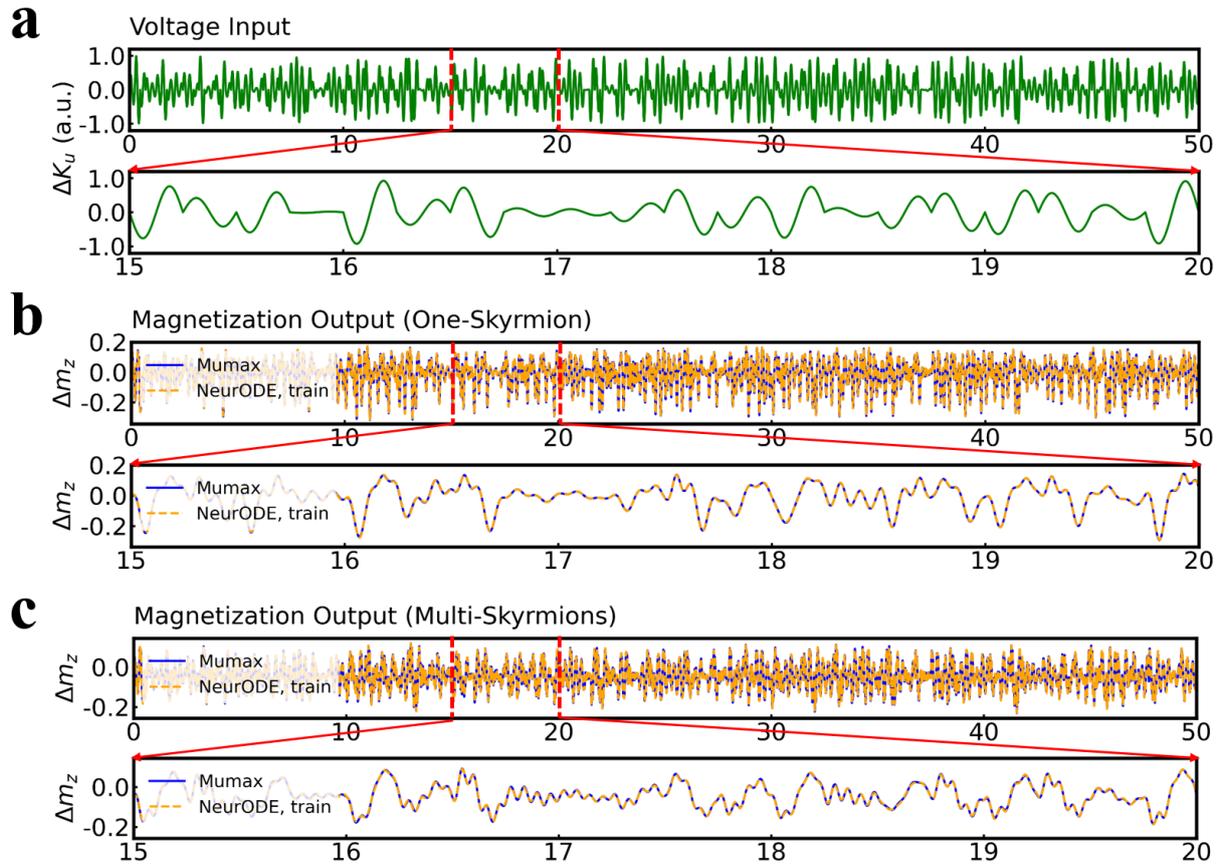

**Supplementary Figure 1. Training performance of Neural ODEs for the skyrmion systems using voltage as input. a** Normalized random sine voltage input, in the form of $\Delta K_u$, with amplitude ranging from $-2\times10^5$J/m$^3$ to $2\times10^5$J/m$^3$ (corresponding to -1 to 1 in the graph). Here, the variation of the PMA ($\Delta K_u$) is linearly dependent on the voltage applied, because of the VCMA effect. Comparisons of the predicted training output of $\Delta m_z$ by Neural ODE (orange dashed curve) and micromagnetic simulation output (blue curve) for the one-skyrmion system without grain inhomogeneity (**b**) and the multi-skyrmions system with grain inhomogeneity in (**c**).

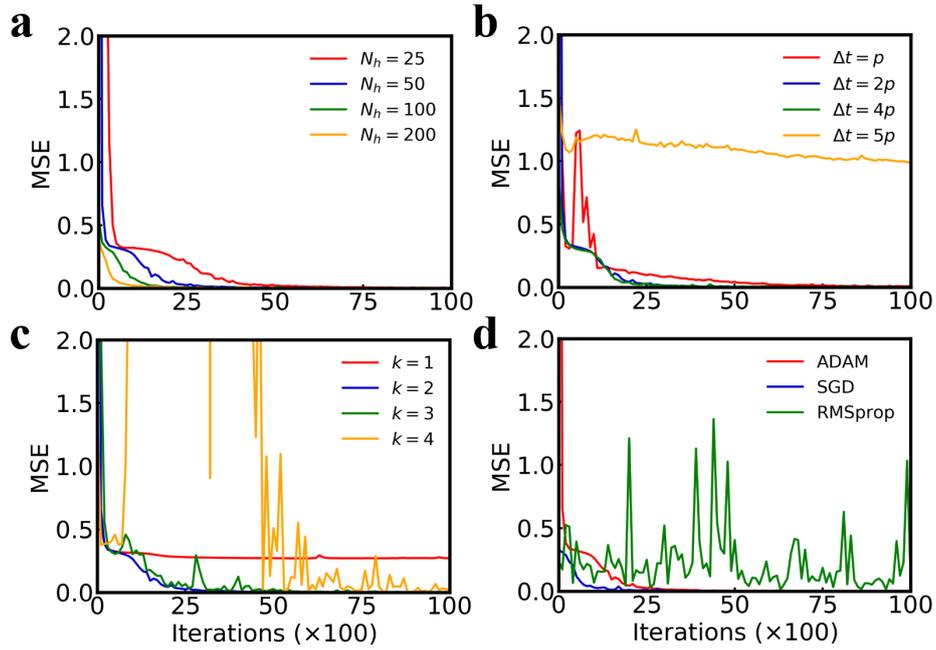

**Supplementary Figure 2. Training performance of Neural ODE.** Training loss (MSE) of the one-skyrmion system without grain inhomogeneity as a function of iterations, for different numbers of units $N_h$ in hidden layers (**a**), sampling intervals $\Delta t$ of the initially observed trajectory $y_1$ (**b**), dimensions of the Neural ODE $k$ (with $k-1$ being the number of delays) (**c**), and optimization algorithms of training (**d**).

## Supplementary Note 2: Training Neural ODE models using incomplete noisy data of skyrmion dynamics

This note investigates the training of Neural ODEs over noisy data. We use the simulation data of the one-skyrmion system as a demonstration. We obtain noisy training data by artificially adding random Gaussian noise with mean $\mu = 0$ and standard deviation $\sigma = 0.1$ into the normalized training data ($\Delta m_z$), as shown in Fig. S3a. Fig. S3c presents the training loss (MSE) as a function of iterations, obtained when training Neural ODEs of dimension $k = 2, 3, 4, 5$ over this noisy data. These results differ from the training results obtained in the noiseless system (see Fig. S2c), where $k \geq 2$ guarantees a sufficiently good model to be trained. Here, the larger the $k$, the better the model is. Further, we remark that the fitted standard deviations $\sigma$ of the training error distribution (see Fig. S3b) of the trained Neural ODEs, compared to the target trajectories, gradually approach 0.1 when $k$ is increased. This 0.1 value corresponds to the standard deviation of the added noise. Fig. S3d shows the final training results of the Neural ODE for different $k$ values.

The Neural ODEs used in this note featured $N_h = 50$ units per hidden layer.

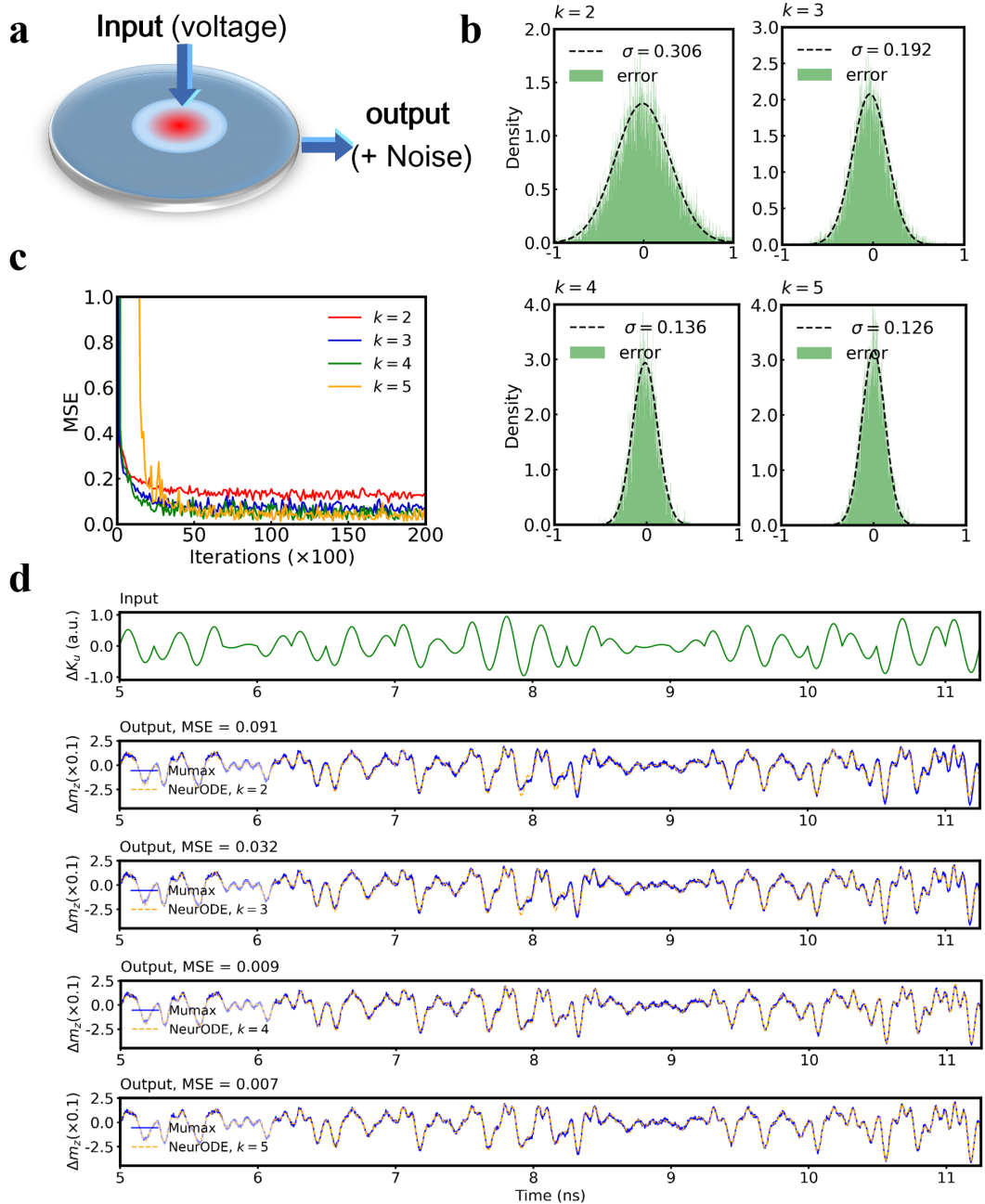

**Supplementary Figure 3. Training performance of Neural ODE model using incomplete noisy data.** **a** Schematic graph of the procedure to obtain training time series with noise, by artificially adding random values drawn from a Gaussian distribution with mean $\mu = 0$ and standard variance $\sigma = 0.1$ into the output time series of the normalized $\Delta m_z$. **b** Training error distribution of the trained Neural ODE compared to the target trajectory. **c** Training loss (MSE) as a function of iterations. **d** Final predicted training output of Neural ODE. All results are presented for different Neural ODE dimensions $k$.

**Supplementary Note 3: Mackey-Glass time-series prediction task evaluation**

This note provides additional results on the skyrmion-based Mackey-Glass time series prediction task. To perform this task, we send the preprocessed input data into the reservoir, simulated with the trained Neural ODE or Mumax (as a control). With $N_r = 50$ virtual nodes in the reservoir, the total number of input values, including both training and testing set is $50 \times 10,000$. Each value is fed into the reservoir for a duration of $t_{step} = 2p$ ($p$ = 2.5 ps). The output trajectory comparison is shown in Fig. S4a and b for the one-skyrmion system without grain inhomogeneity and multi-skyrmions system with grain inhomogeneity, respectively.

Using the trained model, we evaluated the reservoir performance with different $N_r$ values, as shown in Fig. S4c. These results show that the prediction capability of the reservoir is not improved by increasing the number of virtual nodes of the reservoir. On the other hand, Fig. S4d shows that the prediction accuracy can be relatively improved by increasing the step time $t_{step}$.

To improve the reservoir performance further, we also proposed a new method for output reconstruction. Instead of using only the reservoir's output from the current step, we concatenate the response matrix $A$ at each column with output states from previous steps to form a new matrix $A^* \in \mathbf{R}^{(N_r \cdot (1+n_p)) \times T_r}$, where $n_p$ is the number of previous steps' states. In this way, the information of the reservoir states for training are enriched through compensation of previous states from reservoir. Better accuracy of prediction can then be achieved, as shown in Fig. S4e in the case $N_r = 50$. As a final illustration, Fig. S4f also shows selected training set (orange) and testing set (green dashed) for predictions for $H = 20$ and $n_p = 16$ (the red dot in Fig. S4e) compared to the true trajectory (blue) of Mackey-Glass time series, using the one-skyrmion system as reservoir, simulated by Mumax and the trained Neural ODE.

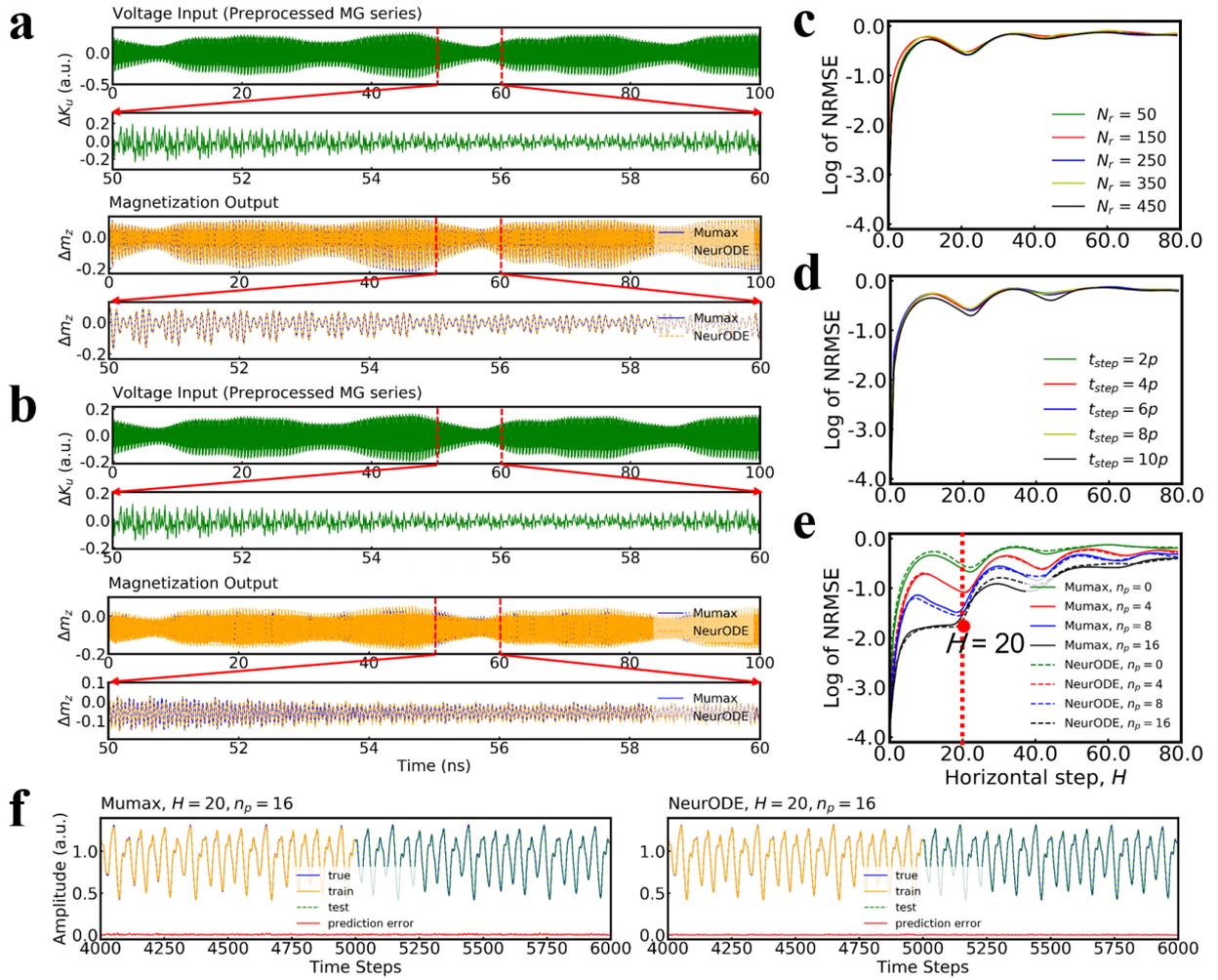

**Supplementary Figure 4. Evaluation of the Mackey-Glass time-series prediction task.** Selected reservoir output states, obtained continuously in time, computed by using Mumax and the trained Neural ODE, with preprocessed Mackey-Glass time series as input. The results are presented for the one-skyrmion system without grain inhomogeneity in **a** and for the multi-skyrmions system with grain inhomogeneity in **b**. Prediction accuracy in terms of NRMSE as a function of predicted horizontal steps for different **c** $N_r$ values, **d** time durations $t_{step}$ for each preprocessed input value fed into the reservoir, and **e** numbers of considered previous steps' states $n_p$. **f** Selected training set (orange) and testing set (green, dashed) for Mackey-Glass predictions for $H=20$ and $n_p = 16$ (red dot in Fig. S1**e**) compared to the true trajectory (blue) of Mackey-Glass time series using one skyrmion system as reservoir, simulated by Mumax and Neural ODE.

## Supplementary Note 4: Modeling of the skyrmion system with electric current as input

The main article focuses on the voltage control of skyrmion-based devices. Skyrmions can also be moved by an electric current through current-induced spin–orbit torques (see Fig. S5a). Skyrmions are easily deformed when they are moving in a confined nanodisk, making this situation quite complex. NeurODE are highly attractive for modeling this situation, and we demonstrate here that the trajectory of a single skyrmion (position of the skyrmion core $x_c, y_c$) as well as the resulting mean magnetization of the device can be modeled using a Neural ODE.

The training dataset is again obtained from Mumax simulations. We use as input a random sine electric current, with amplitude of current density ranging from $-1 \times 10^{11}$ to $1 \times 10^{11}$ A/m$^2$, and a frequency $f = 0.5$ GHz (see Fig. S5c). The output trajectory of the skyrmion is recorded every $p = 20$ ps. A three-layer neural network with 100 units in each hidden layer is trained using 40,000 position data points. The final training output, as well as its comparison with the Mumax results, are shown in Fig. S5c. To further infer the magnetization information, another three-layer neural network $M_\theta^{out}$ with 50 units in each hidden layer was trained to recognize the variation of mean perpendicular magnetization $\Delta m_z$ with skyrmion positions as input into the neural network (see Fig. S5b).

The trained Neural ODE is then evaluated on the task of Mackey-Glass time-series prediction, as in the voltage-control case. The preprocessed Mackey-Glass series, with $N_r = 50$, in the form of an electric current, is sent into Mumax and Neural ODE, correspondingly, with $t_{step} = 8p$. We obtained 410,000 data points (corresponding to 8,200 points of Mackey-Glass series) of final output trajectory after a simulation time of more than 40 days for Mumax and 80 minutes for Neural ODE. The corresponding magnetization $\Delta m_z$ is then calculated by sending the output from Neural ODE into $M_\theta^{out}$. MSE of $\Delta m_z$ over the whole 410,000 data points computed by Neural ODE and $M_\theta^{out}$ is as low as $7.09 \times 10^{-5}$ compared to that simulated by Mumax (see Fig. S6a).

The Mackey-Glass time series prediction error (NRMSE) as a function of predicted horizontal step by Mumax and Neural ODE is shown in Fig. S6b. Fig. S6c shows the selected comparative prediction results of $H = 1$ and $H = 25$ (red dots in b) obtained by Mumax and Neural ODE model respectively. Excellent agreement is again seen. Here, we used the first $T_r = 5,000$ points for training and the rest for testing.

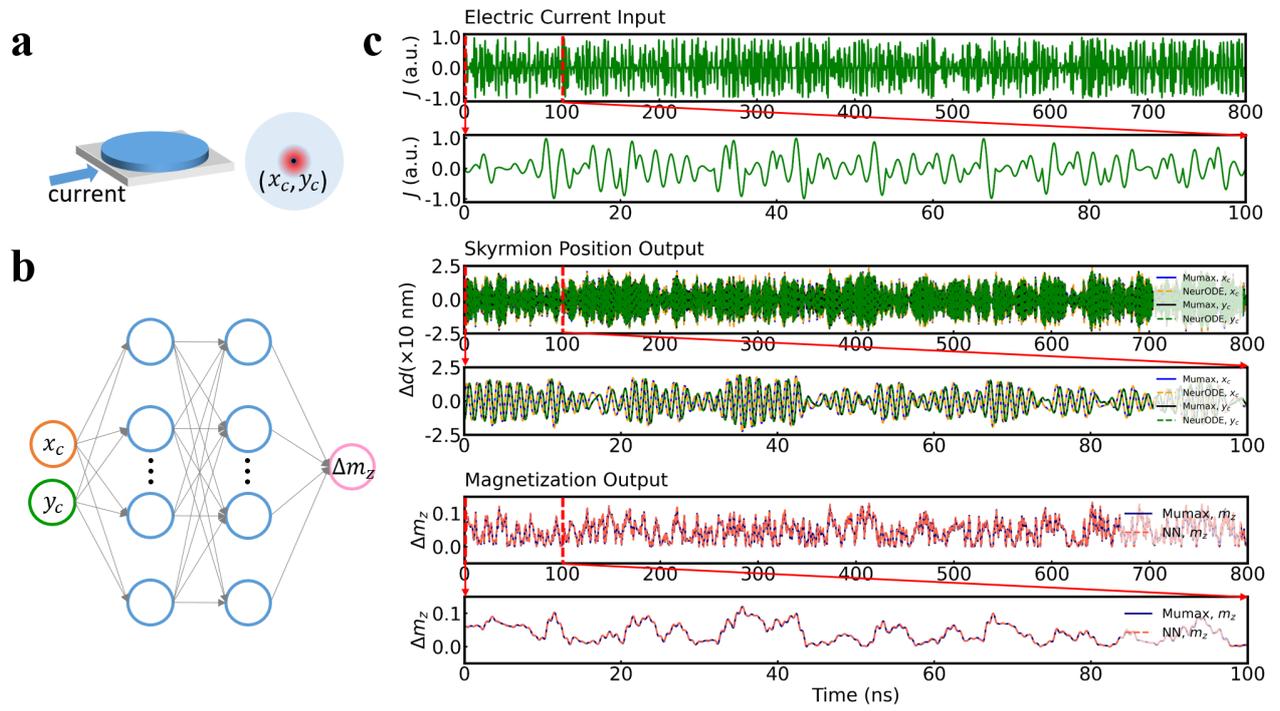

**Supplementary Figure 5. Modeling of the skyrmion system with electric current as input. a** Schematic of the nanodisk where skyrmion exists in the magnetic layer (blue), and electric current applied through the heavy metal layer (grey) is used as an external input. **b** Topology of the neural network used for obtaining the mean perpendicular magnetization $\Delta m_z$ from the skyrmion core position $(x_c, y_c)$. **c** Random sine electric current input, with amplitude ranging from -1×10$^{11}$ to 1×10$^{11}$ A/m$^2$ and frequency $f = 0.5$ GHz, into the skyrmion system, and corresponding output trajectory of skyrmion position $(x_c, y_c)$ calculated by Mumax and by the trained Neural ODE, as well as $m_z$, obtained from Neural ODE results and from Mumax.

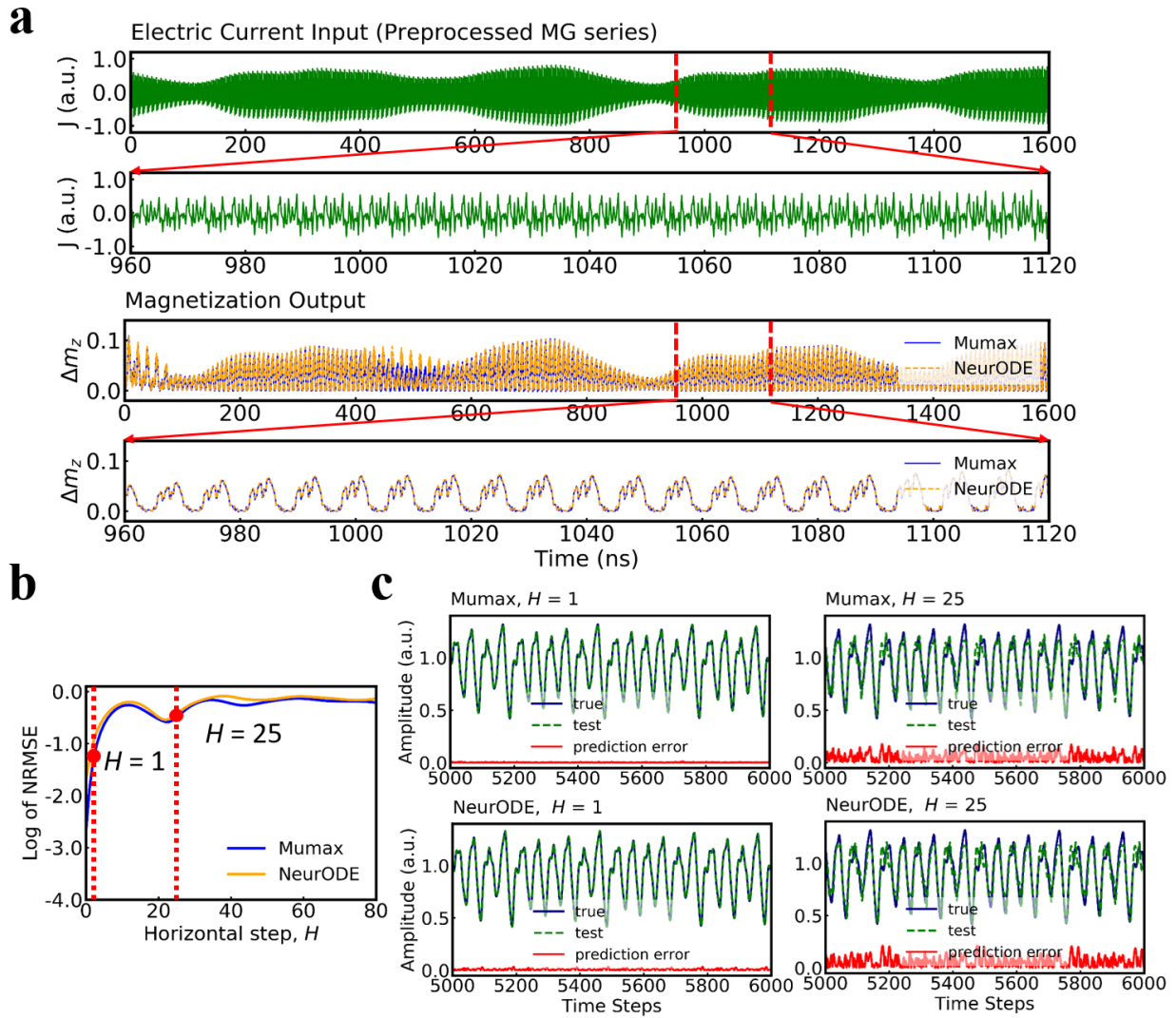

**Supplementary Figure 6. Performance of Mackey-Glass time-series prediction, using Neural ODE and Mumax, for the skyrmion system with electric current input. a** Selected reservoir output states, continuously in time, computed using Mumax and the trained Neural ODE with preprocessed Mackey-Glass time series as input. **b** Mackey-Glass time-series prediction error (NRMSE) as a function of predicted horizontal step by Mumax and Neural ODE. **c** Selected training set (orange) and testing set (green dashed) trajectories for Mackey-Glass predictions for $H = 1$ (short term prediction) and $H = 25$ (long term prediction) (red dots in **b**), compared to the true trajectory (blue) of Mackey-Glass time series, using the one-skyrmion system as reservoir by Mumax and Neural ODE.

## Supplementary Note 5: Spoken digit recognition performance of Neural ODE based on the experimental measurement of spintronic oscillator

This note provides additional results on the use of Neural ODEs to predict experimental measurements, on the spoken digit recognition task with spintronic oscillator. Fig. S7 presents the same results as Fig. 4 in the main body text for the spectrogram method (Fig. 4 in the main body text was presented with the cochlear method, see Methods section). Overall, the results appear highly similar to the cochlear method case. A Neural ODE is able to fit the training set very accurately if its dimension is higher than two (Figs. S7a-b). The trained neural ODE is able to predict the output signals of unseen experiments precisely (Fig. S7c). Finally, with the addition of noise (Fig. S7d), a Neural ODE is able to predict the accuracy of the experiment of the task of spoken digit recognition perfectly (Fig. S7d).

A surprising result is that, with the addition of noise, the spoken digit task performance degrades for the spectrogram method, while the task performance improved for the cochlear method (see Fig. 4, main body text). These results suggest that the output states from the reservoir are highly over-fitted in the noiseless cochlear case; noise suppresses this over-fitting effect. Conversely, in the spectrogram case, the reservoir computer operates in a non-over-fitted regime; noise then plays a more conventional detrimental role. The capability of Neural ODEs to capture such subtle behaviors showcases their impressive predictive power.

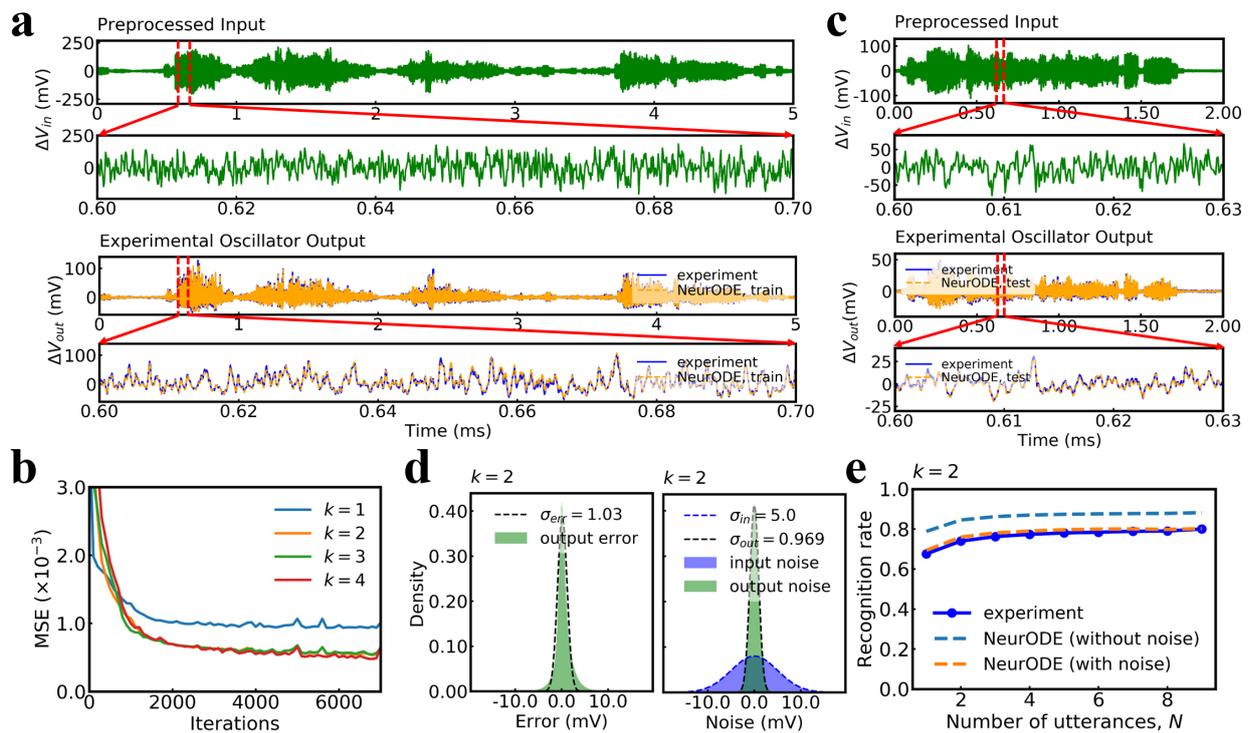

**Supplementary Figure 7. Prediction of experimental results using Neural ODEs (spectrogram method). a** Training output trajectory of voltage $\Delta V_{out}$ by Neural ODE (dashed orange) with corresponding preprocessed input $\Delta V_{in}$, in comparison with the experimental measurement (blue), for $k = 2$. This training set is adopted from the first utterance of the first speaker. A three-layer neural network $f_\theta$ with each hidden layer of 100 units is trained. **b** Training loss (MSE) of Neural ODE with $k = 1, 2, 3, 4$ as a function of iterations. **c** Selected response output trajectories predicted by the trained Neural ODE with corresponding preprocessed input, in comparison with the experimental output, for digit nine of the seventh utterance of the fifth speaker. **d** Left to right: Error distribution (green shadow) and fitted Gaussian distribution (dashed curve) extracted by computing the difference between the output predicted by Neural ODE and the experimental measurement, a Gaussian noise (purple shadow) added in the preprocessed input into the Neural ODE and the corresponding noise distribution (green shadow) with fitted pdf (black dashed curve) in the predicted output trajectory solved by Neural ODE. $\sigma_{in}$ was adjusted so that $\sigma_{out} = 1.03$ mV is close to $\sigma_{err} = 0.97$ mV. **e** Spoken digit recognition rate in the testing set as a function of utterances $N$ used for training. The solid curve, blue dashed curve, orange dashed curve are the experimental result, Neural ODE result with noise considered, Neural ODE result without any noise, respectively.

## Supplementary Note 6: Mathematical derivations

### A Converting a system of ODEs into a single higher-order ODE in one variable

**Theorem 1 (for linear system)** Let $A$ be an $n \times n$ matrix, and let $p_A(\lambda)$ be the characteristic polynomial of $A$. Let $\mathbf{y}(t) = (y_1(t), \ldots, y_n(t))^T$ be a solution to the system of ODEs

$$\dot{\mathbf{y}} = A\mathbf{y}. \tag{1}$$

Then, each coordinate function $y_j(t)$ satisfies the $n$th-order differential equation

$$p_A(D)y_j = 0. \tag{2}$$

*Proof*: Rewrite the differential equation using the operator notation as

$$D\mathbf{y} = A\mathbf{y}, \tag{3}$$

where $D\mathbf{y}$ indicates differentiation of the vector $\mathbf{y}$ by $\frac{d}{dt}$ and $A\mathbf{y}$ indicates multiplication of vector $\mathbf{y}$ by the matrix $A$. By differentiating both sides of the ODEs, we obtain

$$\begin{aligned} D^2\mathbf{y} &= D(A\mathbf{y}) = A(D\mathbf{y}) = A^2\mathbf{y} \\ D^3\mathbf{y} &= A^3\mathbf{y} \\ &\vdots \\ D^n\mathbf{y} &= A^n\mathbf{y}. \end{aligned} \tag{4}$$

Hence, $p(D)\mathbf{y} = p(A)\mathbf{y}$ for any polynomial $p(\lambda)$ by linearity. The Cayley-Hamilton theorem states that $p_A(A) = 0$. Therefore, $p_A(D)\mathbf{y} = 0$. So, for each coordinates, $p_A(D)y_j = 0$.

**Differential elimination procedure (for nonlinear system).** The procedures of converting systems of ODEs to a single-higher order ODE in one variable was mathematically proved for nonlinear systems, we refer readers to refs[1–3] for further details. Here, we are giving the general ideas of the procedures followed in these references. We consider a system of nonlinear ODEs of the form ($\dot{\mathbf{x}} = f(\mathbf{x})$):

$$\begin{aligned} \dot{x}_1 &= f_1(x_1, x_2, \ldots, x_n) \\ \dot{x}_2 &= f_2(x_1, x_2, \ldots, x_n) \\ &\vdots \\ \dot{x}_n &= f_n(x_1, x_2, \ldots, x_n) \\ y &= x_j, j \in (1, \ldots, n). \end{aligned} \tag{5}$$

Here $\mathbf{x} = (x_1, x_2, \ldots, x_n)$ is an n-dimensional state variable, and $y$ is considered as an output vector, which could be any component of the state variable. We assume that $f(\mathbf{x})$ is rational polynomial functions of the state variables, a reasonable assumption in most applications. Our aim is to obtain a single ODE in the variable $y$.

The procedure can be implemented by taking a sufficient number of derivatives of the system, followed by computation of a Gröbner Basis of the new system[1–3]. In general, for elimination to work, the number of equations must be strictly greater than the number of unknowns[1,2]. Firstly, from the output equation, we can obtain $\dot{y} = \dot{x}_j$ and $\ddot{y} = \ddot{x}_j$. This, however, introduces the second derivative of $x_j$, and thus differentiation of the corresponding state variable equation is needed:

$$\begin{aligned}
\dot{x}_1 &= f_1(x_1, x_2, \ldots, x_n) \\
\dot{x}_2 &= f_2(x_1, x_2, \ldots, x_n) \\
&\vdots \\
\dot{x}_n &= f_n(x_1, x_2, \ldots, x_n) \\
y &= x_j \\
\dot{y} &= \dot{x}_j \\
\ddot{y} &= \ddot{x}_j \\
\ddot{x}_j &= \sum_{i=1}^{n} \frac{\partial f_i}{\partial x_i} \dot{x}_i, j \in (1, \ldots, n),
\end{aligned} \quad (6)$$

Now, at the second step, we have $n+4$ equations and $2n+1$ unknowns $(x_1, x_2, \ldots, x_n, \dot{x}_1, \dot{x}_2, \ldots, \dot{x}_n, \ddot{x}_j)$. So, for $n > 2$, the number of unknowns is greater than or equal to the number of equations. Next, at step three, we take an additional derivative of the output equation, which causes the third derivative of the state variable to appear. Thus, we take the corresponding derivative of the state variable $x_j$ and add the following equations to the system:

$$\begin{aligned}
\dddot{y} &= \dddot{x}_j \\
\dddot{x}_j &= \sum_{i=1}^{n} \left[\frac{\partial^2 f_j}{\partial x_i \partial t} \dot{x}_i + \frac{\partial f_j}{\partial x_i} \ddot{x}_i\right], j \in (1, \ldots, n) \\
\ddot{x}_k &= \sum_{i=1}^{n} \frac{\partial f_k}{\partial x_i} \dot{x}_i, \text{ for } k = 1, \ldots, n \text{ and } k \neq j.
\end{aligned} \quad (7)$$

In doing this, we add $2 + n - 1 = n + 1$ equations and $1 + (n-1) = n$ unknowns; so, we have $2n+5$ equations and $3n+1$ unknowns. So, for $n > 3$, the number of unknowns is greater than or equal to the number of equations. Differentiation of the output equation and corresponding state variable equations is continued until the number of equations is greater than the number of unknowns.

**Lemma 1** Let the system consist of $n$ state variables. There is always an equal number of equations and unknowns when there are $n-1$ derivatives of the output equation.

*Proof*: Initially, there are $n+1$ equations and $2n$ unknowns. After one step derivative on the output, there are always $n+2$ equations and $2n$ unknowns. Thus, there are $n-2$ more unknowns than equations. In each step thereafter, we add one more equation than unknown, so at the $(n-1)$th step ($(n-1)$th-order derivatives to the output), there are zero more unknowns than equations. Thus, the Lemma is proven.

**Corollary to Lemma 1** At the $n$-th step, there is one more equation than unknown.

This suggests that an ODE of the same (or possibly lower) differential order as the number of state variables can always be obtained. In other words, since there is one more equation than unknown at step n, an ODE of differential order $n$, or lower, can be obtained.

**Remark**: If there is a time-dependent input $e$ added into the system, we use the same elimination procedure.

Then, after taking the second derivative of the output y, we have the following $n+4$ equations

$$\begin{aligned}
\dot{x}_1 &= f_1(x_1, x_2, \ldots, x_n, e) \\
\dot{x}_2 &= f_2(x_1, x_2, \ldots, x_n, e) \\
&\vdots \\
\dot{x}_n &= f_n(x_1, x_2, \ldots, x_n, e) \\
y &= x_j \\
\dot{y} &= \dot{x}_j \\
\ddot{y} &= \ddot{x}_j \\
\ddot{x}_j &= \sum_{i=1}^{n} \frac{\partial f_i}{\partial x_i}\dot{x}_i + \frac{\partial f_j}{\partial e}\dot{e}, j \in (1, \ldots, n),
\end{aligned} \quad (8)$$

in which the first derivative of $e$ is included. Thus, after taking $n$th-order derivatives to the output $y$ at $n$-th step, the first to $(n-1)$th-order derivative of $e$ are included, and a single $n$th-order ODE in variable $y$ can be obtained, which explains why the time-delayed state of input $e(t)$ should be included when training the neural network.

## B Noise level for higher-order derivatives

We now suppose the observed data $s(t) = y(t) + n(t)$ is a stationary random process[4], recorded with measurement noise $n(t)$, identically distributed with variance $\sigma_{noi}^2$, zero mean, and a normalized autocorrelation function $c_{noi}(\tau) = E[n(t)n(t-\tau)]/\sigma_{noi}^2$. Let $y(t)$ be the clean variable with variance $\sigma_{sig}^2 = E[y(t)^2] - E[y(t)]^2$ and normalized autocorrelation function $c_{sig}(\tau) = E[(y(t) - E[y(t)])(y(t-\tau) - E[y(t)])]/\sigma_{sig}^2$. We assume that the data is recorded with a high sampling rate $\Delta t$, such that the successive observations are strongly correlated. If we take the first derivative through the following form

$$\dot{s}(t) = \frac{1}{2\Delta t}(s(t + \Delta t) - s(t - \Delta t)) \quad (9)$$

The real signal derivative is, neglecting terms higher than the second order in $\Delta t$,

$$\dot{y}(t) = \frac{1}{2\Delta t}(y(t + \Delta t) - y(t - \Delta t)) \quad (10)$$

Therefore the corresponding (derivative) signal variance is

$$\begin{aligned}
\sigma_{sig,der}^2 &= E[\dot{y}(t)^2] - E[\dot{y}(t)]^2 \\
&= E[\dot{y}(t)^2] \\
&= E[\frac{1}{4\Delta t^2}(y(t+\Delta t) - y(t-\Delta t))^2] \\
&= \frac{1}{2\Delta t^2}(E[y(t)^2] - E[y(t-\Delta t)y(t+\Delta t)]) \\
&= \frac{1}{2\Delta t^2}(1 - c_{sig}(2\Delta t))
\end{aligned} \quad (11)$$

The corresponding noise of the derivative signal $\dot{s}(t) - \dot{y}(t)$ is

$$\sigma_{noi,der}^2 = \frac{1}{2\Delta t^2}(1 - c_{noi}(2\Delta t)) \quad (12)$$

Thus, the noise to signal ratio of the second derivative is

$$\frac{\sigma_{noi,der}}{\sigma_{sig,der}} = \frac{\sigma_{noi}}{\sigma_{sig}} \sqrt{\frac{1-c_{noi}(2\Delta t)}{1-c_{sig}(2\Delta t)}}. \tag{13}$$

For white noise, the autocorrelation function is an impulse at lag 0, thus $c_{noi}(2\Delta t)$ is close to zero. Therefore, the noise-to-signal ratio of the derivative ($\sigma_{noi,der}/\sigma_{sig,der}$) can be much larger than the original signal ($\sigma_{noi}/\sigma_{sig}$) if the sampled adjacent signal is strongly correlated, which is a common situation in most applications. This consideration explains our choice, in the main paper, to use delayed signals instead of derivates as state variables for Neural ODEs for systems with incomplete information of dynamics.

## Supplementary Note 7: Alternative interpretation based on the embedding theorem

An alternative interpretation for our method can also be obtained through the "embedding theorem" for state-space reconstruction of dynamical systems. It provides additional insight into our technique. Let us assume the system can be described by an ensemble of physical variables represented by vector **u**, and that the variable that can we can measure (e.g., a voltage) can be computed by a measurement function $s(\mathbf{u})$. Let us assume that the underlying ODE describing the physics of the system is $\dot{\mathbf{u}} = F(\mathbf{u},t)$. Using the point of view of the embedding theorem, our technique corresponds to changing the physical variables **u** to new variables **y**. We have $y_1 = s(\mathbf{u}(t))$ and $y_2 = y_1(t + \Delta t_d) = s(\mathbf{u}(t) + \int_t^{t+\Delta t_d} F(\mathbf{u},t')\,dt')$. We define the transformation $G$ by $G(\mathbf{u}(t)) = \mathbf{u}(t) + \int_t^{t+\Delta t_d} F(\mathbf{u},t')\,dt'$ so that we have $y_2 = s(G(\mathbf{u}(t)))$ and, for all $i$ values,

$$y_i = y_1(t + i\Delta t_d) = s(G^i(\mathbf{u}(t))). \tag{14}$$

According to the embedding theorem[4-7], for almost every smooth measurement function $s$ and positive $\Delta t_d$ value, such a delay map $\mathbf{y}(t)$ is a smooth, one-to-one coordinate transformation with a smooth inverse if its dimension $k$ is at least twice the capacity dimension of the dynamical system (also called fractal dimension or box-counting dimension). This consideration provides an alternative justification for the choice of these coordinates, in the main paper.